\documentclass{article}

\PassOptionsToPackage{numbers, compress}{natbib}

\usepackage[preprint]{neurips_2026}

\usepackage[utf8]{inputenc}
\usepackage[T1]{fontenc}
\usepackage{hyperref}
\usepackage{url}
\usepackage{graphicx}
\usepackage{subcaption}
\usepackage{booktabs}
\usepackage{microtype}
\usepackage{float}            

\usepackage{amsmath}
\usepackage{amssymb}
\usepackage{amsfonts}
\usepackage{mathtools}
\usepackage{amsthm}
\usepackage{nicefrac}

\usepackage[capitalize,noabbrev]{cleveref}

\usepackage{multicol}
\usepackage{multirow}
\usepackage{makecell}
\usepackage{subcaption}

\newcommand{\figref}[1]{Fig.~\ref{#1}}

\newcommand{\methodname}{VideoMDM}

\makeatletter
\renewcommand{\paragraph}[1]{\par\vspace{-\parskip}\noindent\textbf{#1}}

\usepackage{xcolor} 
\usepackage{colortbl}
\usepackage{soul}
\definecolor{tabfirst}{RGB}{255, 165, 165}   
\definecolor{tabsecond}{RGB}{255, 210, 165}  
\definecolor{tabthird}{RGB}{255, 255, 165}   

 \newcommand{\amir}[1]{{#1}}
 \newcommand{\amdel}[1]{}
 \newcommand{\gal}[1]{{#1}}
 \newcommand{\galdel}[1]{}
 \newcommand{\orl}[1]{}
 \newcommand{\merav}[1]{}

\title{VideoMDM: Towards 3D Human Motion Generation From 2D Supervision}

\author{%
  \textbf{Amir Mann}$^1$ \quad
  \textbf{Gal Michael Harari}$^1$ \quad
  \textbf{Merav Keidar}$^1$ \quad
  \textbf{Or Litany}$^{1,2}$ \\[1.5ex]
  $^1$Technion \quad $^2$NVIDIA \\[1.5ex]
  \url{https://videomdm.github.io}
}

\begin{document}

\maketitle

\begin{figure}[H]
  \centering
  \includegraphics[width=\textwidth]{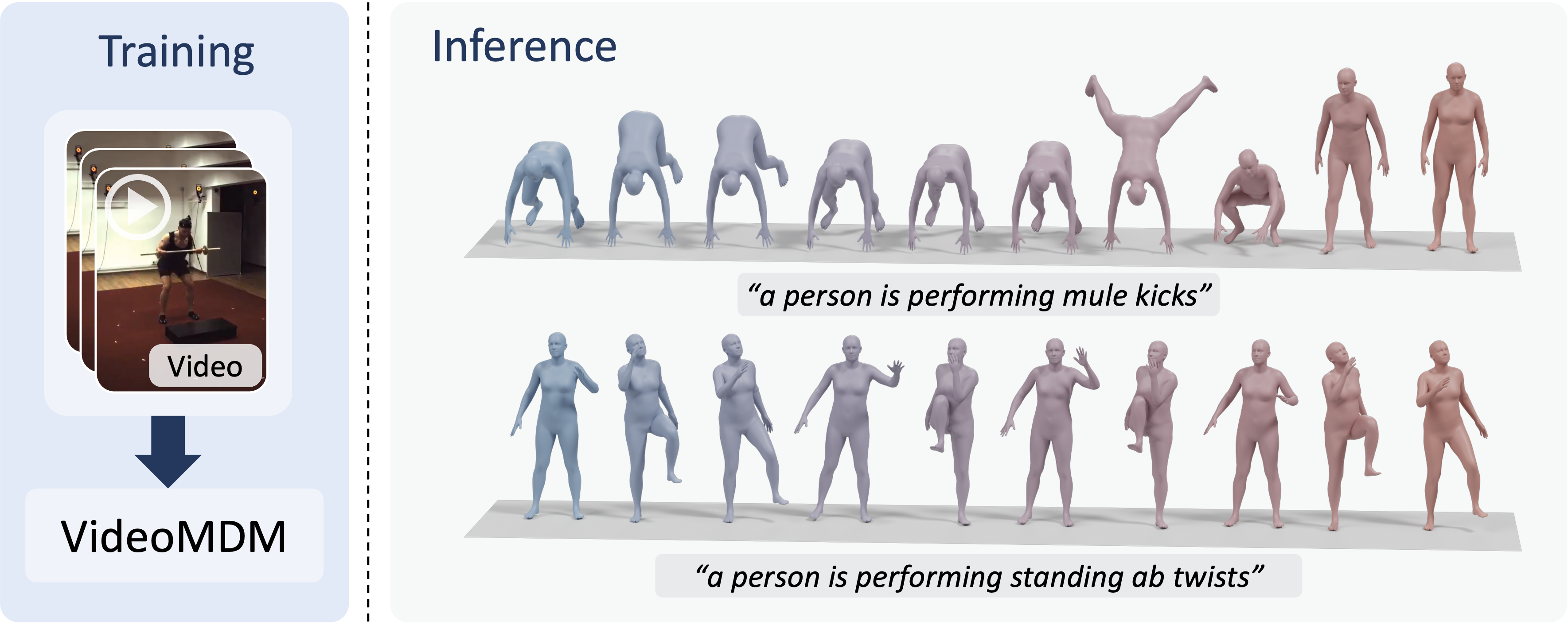}
  \caption{\textbf{We demonstrate VideoMDM on monocular videos of human activities.}
  Our framework trains 3D text-to-motion diffusion models using 2D pose sequences extracted from videos.
  Left: representative training videos. Right: generated motions using the trained model from text prompts.
  Despite relying solely on 2D supervision, VideoMDM attains motion fidelity approaching that of fully 3D-supervised training. See \href{videomdm.github.io}{project page} for animated results, code and more.}
  \label{fig:teaser}
\end{figure}

\begin{abstract}
We introduce VideoMDM, a diffusion-based framework that trains 3D human motion priors directly from accurate 2D poses extracted from monocular videos, without any 3D ground truth. A pretrained 2D-to-3D lifter provides approximate 3D pose sequences that serve as a noisy teacher: these are diffused, denoised by the model in 3D, and supervised in 2D by reprojecting the prediction and comparing against accurate keypoints. We show that, under mild assumptions, a depth-weighted 2D reprojection loss is equivalent in expectation to direct 3D supervision, and we adapt standard 3D motion regularizers — velocity consistency and over-parameterized representation alignment — to this 2D setting. Unlike methods that lift 2D to 3D only at inference, VideoMDM learns a coherent 3D motion manifold during training. On HumanML3D it nearly closes the gap to fully 3D-supervised MDM (FID 0.88 vs 0.54); On real video datasets Fit3D and NBA the method learns to generate motions consistently preferred by humans, with strong quantitative results.
\end{abstract}
\section{Introduction}
\label{sec:intro}

Generating realistic 3D human motion is central to animation, gaming, simulation, and embodied AI. Diffusion-based motion models such as MDM \cite{tevet2023human} have recently achieved striking realism when trained on motion-capture (MoCap) data.  
Yet, their success remains tightly coupled to the availability of high-quality 3D supervision: MoCap datasets are captured in controlled studio environments and span only a narrow subset of real-world movement. Models trained on them inherit limited diversity and fail to capture the richness of human motion observed in the wild.

At the same time, vast amounts of online video depict human actions across diverse environments, identities, and viewpoints.  
Harnessing such in-the-wild data could enable scalable and diverse 3D motion generation.  
However, most videos are monocular, lacking the multi-view cues necessary for reliable 3D reconstruction.  
While monocular 3D pose and motion estimators \cite{motionbert2022,wham:cvpr:2024,elepose2022cvpr,li2025mvlift} remain noisy and ambiguous, 2D keypoint detectors \cite{cao2019openposerealtimemultiperson2d,10.1109/TPAMI.2022.3222784,jiang2023rtmposerealtimemultipersonpose} have reached high accuracy and robustness.  
The key challenge, therefore, is how to train a generative 3D motion model using only accurate 2D supervision derived from monocular video.  

We address this challenge by introducing \methodname{}, a diffusion-based framework for training 3D human motion models entirely from 2D pose supervision.  
Unlike prior approaches that \amir{triangulate} 2D motions to 3D only at inference time \cite{kapon2024mas, li2025mvlift}, which prevents the model from learning a consistent 3D prior, or that depend on 3D supervision for fine-tuning \cite{Guo_2025_ICCV}, \methodname{} trains a diffusion model natively in 3D space using only 2D supervision. Our formulation opens, for the first time, a path towards large-scale training of \amir{3D} text-to-motion diffusion from monocular videos without any MoCap data.

Building on cross-modality training of Image-to-3D diffusion~\cite{peng2025lesson}, we adopt a noisy-teacher scheme: a pretrained 2D-to-3D lifter produces approximate 3D pose sequences from 2D inputs; these are diffused, denoised by the model in 3D, and supervised in 2D by reprojecting the prediction and comparing against accurate keypoints from video. This design lets the model learn a coherent 3D motion manifold grounded entirely in 2D observations. 

To make 2D-only diffusion training practical for 3D motion, we introduce a depth-aware reprojection loss that, under mild assumptions on data and camera distribution, is provably equivalent in expectation to standard 3D MSE supervision (\cref{sec:method_depth}, proof in \cref{app:proof}). We further adapt two standard 3D motion regularizers to the 2D setting: a depth-weighted 2D velocity loss for temporal coherence, and a motion representation alignment loss that supervises the over-parameterized motion channels — joint rotations, joint velocities, and foot contacts — via ray-projection pseudo-targets, since no 3D ground truth is available for them \amir{(\cref{sec:method_repr})}.


We evaluate VideoMDM in three regimes. (i) On a 2D-only version of HumanML3D \cite{Guo_2022_CVPR}, where 2D poses are obtained by projecting MoCap, \methodname{} achieves FID 0.88 — nearly closing the gap to fully 3D-supervised MDM (FID 0.54) and improving on the strongest 2D-supervised baseline by roughly x2. (ii) On Fit3D \cite{Fieraru_2021_CVPR} — a real-world setting, where training uses monocular fitness video with extracted 2D keypoints and no 3D supervision — VideoMDM halves joint error against WHAM \cite{wham:cvpr:2024} on motions far outside the lifter's distribution (MPJPE 111 vs 228 mm) and produces ~5.5× smoother motion (Accel 3.16 vs 17.66 m/s²). \amir{And is preferred by humans in generation against all baselines.} (iii) On the NBA dataset \cite{kapon2024mas}, VideoMDM is preferred over MAS in 64\% of pairwise human comparisons.
Together, these results show that 2D supervision alone is sufficient to learn 3D motion priors that are coherent, perceptually realistic, and capable of generalizing beyond the lifter that bootstrapped them. 


Our main contributions are:
\begin{enumerate}
    \item \textbf{The first 2D-supervised diffusion training framework for 3D human motion priors}, enabling high-fidelity prior learning directly from monocular videos without any 3D supervision.
    \item 
    \textbf{A stabilization strategy for condition-free denoising}, leveraging depth-aware weighting and reprojection consistency to keep the 3D denoising dynamics geometrically anchored.
    \item \textbf{A 2D-adapted formulation of strong 3D motion regularizers}, enforcing natural motion through velocity-based and representation-level constraints.
\end{enumerate}

\section{Related Work}
\label{sec:related_work}

\paragraph{Human Motion Generation in 3D.}
Generating human motions in 3D is largely driven by deep neural networks. VAEs \cite{bie2022hitdvaehumanmotiongeneration, Guo_2022_CVPR} were early approaches, while diffusion models \cite{zhang2022motiondiffusetextdrivenhumanmotion, chen2023executing, karunratanakul2023guidedmotiondiffusioncontrollable} such as MDM \cite{tevet2023human} substantially improved fidelity. VQ-VAEs \cite{oord2017vqvae} paired with autoregressive \cite{zou2024parcopartcoordinatingtexttomotionsynthesis} and bidirectional autoregressive \cite{guo2023momaskgenerativemaskedmodeling, pinyoanuntapong2024bammbidirectionalautoregressivemotion, hong2025bipobidirectionalpartialocclusion} models \amir{have established a new state-of-the-art quality}. These approaches typically rely on high-quality 3D motion from MoCap systems, e.g., HumanML3D \cite{Guo_2022_CVPR} built from AMASS \cite{mahmood2019amassarchivemotioncapture} and A2M \cite{guo2020action2motion}, which contains approximately 14 thousand motion sequences.

\paragraph{3D Asset Generation with 2D Priors.}
Generating 3D content from 2D data has been widely explored. Methods leverage strong 2D diffusion priors via score distillation \cite{poole2022dreamfusion, nam2024contrastive, NEURIPS2023_1a87980b, xie2024latte3d} or fine-tune 2D models for novel-view consistency~\cite{liu2023zero, shi2023zero123++, liu2023one, liu2024one, gao2024cat3d, sobol2024zerotoheroenhancingzeroshotnovel} on smaller, curated 3D asset datasets~\cite{2023ObjaverseXL}. Other works train generative models directly in 3D~\cite{alliegro2023polydiffgenerating3dpolygonal, liu2023meshdiffusionscorebasedgenerative3d, 10.1145/3680528.3687699, mu2024gsdviewguidedgaussiansplatting, zeng2022lion}. ``A Lesson in Splat'' \cite{peng2025lesson} formalizes diffusion training for 3D Gaussian Splatting~\cite{Kerbl2023_3DGS} under 2D supervision. A common theme is the reliance on 2D image priors for 3D generation.

\paragraph{2D Pose Extraction from Video.}
2D human pose estimation has become highly reliable. OpenPose \cite{cao2019openposerealtimemultiperson2d} introduced Part Affinity Fields, AlphaPose \cite{10.1109/TPAMI.2022.3222784} improved robustness under occlusion, HRNet \cite{wang2020deephighresolutionrepresentationlearning} preserved high-resolution features, and RTMPose \cite{jiang2023rtmposerealtimemultipersonpose} achieves strong real-time accuracy.

\paragraph{Video-to-3D Pose Extraction.}
Recovering 3D motion from monocular video remains challenging due to depth ambiguity and camera-to-world conversion. WHAM~\cite{wham:cvpr:2024} uses a feed-forward model with image features and 2D poses to infer world coordinates, while COIN \cite{li2024coin} applies 3D motion diffusion inpainting for iterative refinement. Other approaches \cite{wang2024tram, zhang2024rohmrobusthumanmotion} exploit SLAM cues \cite{doi:10.1177/027836498600500404, robotics11010024} \amir{to ground their predictions}. Complementary lines of work focus on lifting from 2D keypoints: MotionBERT \cite{motionbert2022} provides a supervised temporal lifting baseline \amir{trained on 3D} \gal{ data}. \amir{Training only on 2D} \gal{ data,} ElePose \cite{elepose2022cvpr} learns a normalizing-flow prior on 2D poses and uses it to steer a lifting model toward better 3D reconstructions\gal{.} \amir{MVLift \cite{li2025mvlift} employs epipolar-constrained 2D diffusion to create pseudo-3D motions to supervise the training of a final multiview 2D model}. Recent works \cite{zhang2024large, genmo2025} combine generation and pose estimation by jointly training multiple 3D motion-related tasks. However, all of these methods still exhibit a non-negligible error gap with respect to ground-truth motions, even within their training datasets.

\paragraph{3D Human Motion Generation from 2D Data.}
Generating plausible 3D motion 
\gal{supervised}
directly from 2D sequences remains comparatively underexplored. Existing approaches perform inference-time lifting from 2D priors to 3D: MAS \cite{kapon2024mas} performs multi-view ancestral sampling from several 2D motion diffusion models and omits root trajectory. Motion-2-to-3 \cite{Guo_2025_ICCV} trains a 2D diffusion model and then adds consistency layers learned from 3D motion data, relying on 3D supervision to recover root trajectory. These works point to the promise of scaling 2D-centric training while improving world-frame trajectory modeling.
In contrast, we train a 3D-native model that learns motions with any root trajectory solely from 2D supervision, without any 3D ground truth.

\section {Method}

\subsection{Preliminaries: Cross-Modality Diffusion for 3D Generation from 2D Supervision}
\label{sec:background}

Diffusion models are typically trained under a \emph{same-modality assumption}: both the diffused input and its supervision target belong to the same domain.  
In 3D generative modeling, this requires large datasets of ground-truth 3D samples, thus significantly limiting scalability.  
\emph{Lesson in Splats (LIS)}~\cite{peng2025lesson} showed that this constraint can be relaxed: by combining approximate 3D estimates with clean 2D supervision, one can train a 3D denoiser without access to any real 3D data.

Specifically, LIS introduces a weak lifter implemented as a deterministic \emph{2D-to-3D predictor}, which serves as a noisy teacher reconstructing approximate 3D Gaussian-splat scenes from single images.  
For high-noise diffusion timesteps ($t > t^*$), these inaccurate outputs are perturbed with sufficient noise to produce samples statistically aligned with those drawn from the (unknown) clean 3D distribution.  
The model is trained to denoise these samples while supervision is applied in 2D via differentiable rendering.  
At low-noise regimes, LIS employs a multi-step denoising scheme:  
for timesteps $t < t^*$, the input is first further diffused to a higher level $t' > t^*$ and then denoised through a short sequence of \amir{DDIM \cite{song2020denoising}} steps down to $t$, where 2D supervision is applied.  
This strategy ensures the model experiences training samples at low-noise which are critical for generating high-frequency geometric detail\amir{s}.


\subsection{Problem Setup and Formulation}

We are given a collection of monocular human-motion videos, each captured by a static camera. For each video, we extract 2D joint trajectories 
$\mathbf{y} \in \mathbb{R}^{J\times2\times F}$, 
where $J$ denotes the number of joints and $F$ the number of frames.  
Empirically, 2D keypoint extraction is highly accurate and robust~\cite{kapon2024mas}.   
To obtain an approximate 3D signal, we assume access to a pretrained 2D-to-3D lifter $L_\phi$ that produces approximate 3D joint trajectories 
$\tilde{x}_0 = L_\phi(\mathbf{y}) \in \mathbb{R}^{J\times3\times F}$, 
serving as a noisy teacher (\cref{sec:background}).  
Such 2D-to-3D lifting networks are well established in the human-motion literature \cite{motionbert2022, elepose2022cvpr, wham:cvpr:2024}. \amir{When no camera parameters are available they can be estimated by solving PnP on $y$ and $\tilde{x}_0$, see \cref{PNPdetails}.} 

Our objective is to train a 3D generative diffusion model 
$D_\theta(x_t, t, \mathbf{c})$
that can sample realistic 3D motion trajectories, optionally conditioned on a text embedding $\mathbf{c}$, while using only 2D supervision.  
We denote by $\Pi_c(\cdot) = K[R|\mathbf{t}_c](\cdot)$ the camera projection operator, with intrinsics $K$, rotation $R$, and translation $\mathbf{t}_c$, which maps 3D joint coordinates to 2D. We denote by $x$ 3D motion and by $y$ 2D motions, \amir{by} $\hat x_0$ the model predictions, by $x_t$ motions diffused to time $t$ and $x^{(f)}$ the motion at frame $f$.



\subsection{Method Overview}
\label{sec:overview}

\begin{figure*}[h]
    \centering
    \includegraphics[width=\linewidth]{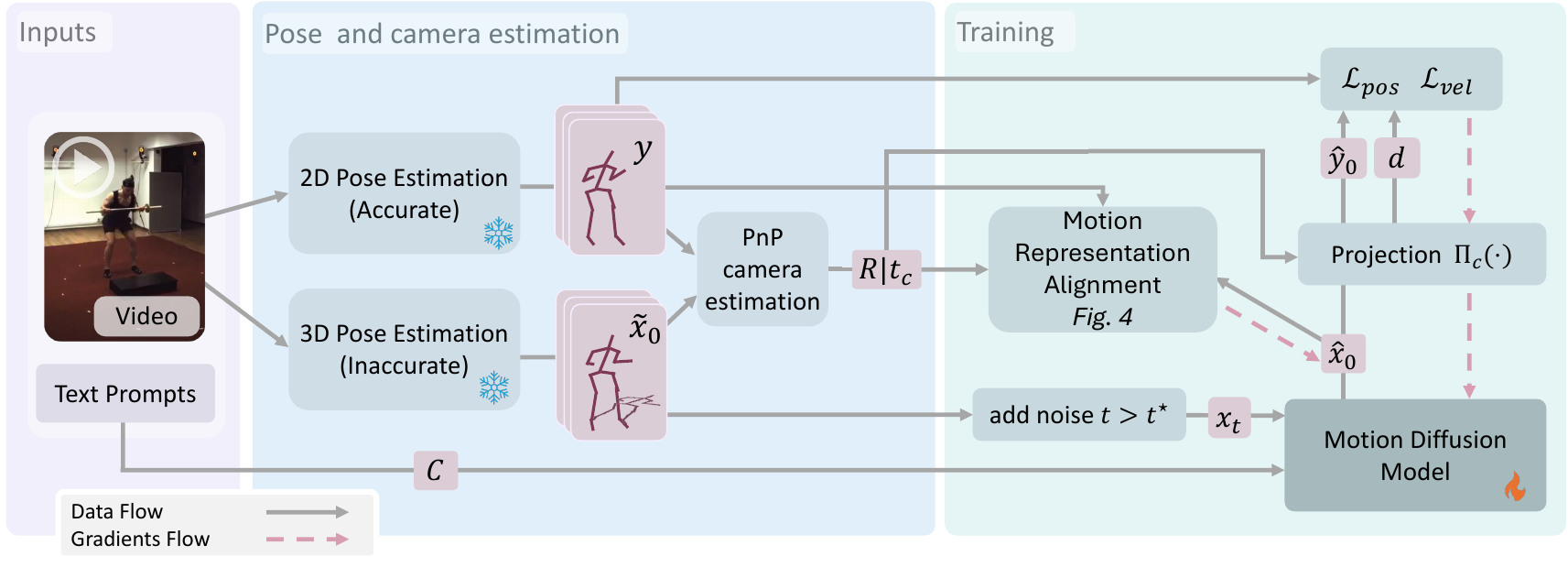}
    \caption{\textbf{VideoMDM training.} From monocular video, we extract accurate 2D keypoints and approximate 3D poses. \amir{A motion diffusion model} is trained to denoise the 3D poses, after diffusing them to a high timestep, under multi-source supervision: (i) 3D representation alignment, and (ii) 2D reprojection and velocity consistency with the accurate 2D pose.}
    \label{fig:Method}
\end{figure*}

The overall framework is illustrated in \figref{fig:Method}.  
As in standard diffusion training we diffuse the lifter’s 3D predictions  $x_t = \sqrt{\alpha_t}\tilde{x}_0 + \sqrt{1-\alpha_t}\boldsymbol{\epsilon}$, where $\boldsymbol{\epsilon}\!\sim\!\mathcal{N}(0,I)$, and train the network to recover the clean 3D motion $\hat{x}_0$~\cite{tevet2023human, ramesh2022hierarchicaltextconditionalimagegeneration}. Crucially, while the denoiser operates in 3D, supervision is applied in the 2D domain through the projection operator $\Pi_c(\cdot)$. 
\amir{With appropriate depth-aware weighting, this 2D objective provides a principled surrogate for 3D supervision as introduced in \cref{sec:method_depth} and proved in \cref{app:proof}}.
To make 2D supervision effective for 
3D motion diffusion, 
several geometric and temporal adaptations are required. \amdel{introwords commented}


\subsection{Depth-Aware Weighting}
\label{sec:method_depth}

A naive 2D reprojection loss is not equivalent to 3D MSE: because perspective projection divides by camera depth $d$, the 2D error is implicitly $1/d$-weighted relative to the underlying 3D error, downweighting distant joints and overweighting near ones. Multiplying the loss by $d$ removes this scaling, produces a loss equivalent in expectation to direct 3D MSE supervision (see \cref{app:proof}). This equivalence holds under two mild assumptions on the data distribution: that the predicted joint depth $d$ matches the true motion depth in the corresponding camera frame, and that training cameras are sampled with uniformly distributed azimuth.


Let $d\in\mathbb{R}^{J\times1\times F}$ denote the predicted depth in the camera coordinate system $\hat{x}_0$. 


We define:
\begin{equation}
\mathcal{L}_{\text{pos}} =
\big\|
d \odot \mathbf{1}_{\{d > d_\text{min}\}} \odot 
\big(\Pi_c(\hat x_0) - y\big)
\big\|_2^2 ,
\end{equation}

\noindent
where $\odot$ denotes element-wise multiplication across joints and frames. The truncation $\mathbf{1}\{d > d_{\min}\}$ drops joints whose predicted depth falls below $d_{\min}$, primarily preventing unreliable gradients when joints are predicted behind or very close to the camera, where the projection equations and the equivalence above no longer hold.

\subsection{Natural Motion Regularization}
\label{sec:method_repr}
\paragraph{2D Velocity Loss.}
Generative motion networks are commonly regularized using additional geometric losses \cite{tevet2023human, Shi_2020}. We adapt the 3D velocity loss to our 2D setup to enforce temporal similarity between the generated motions and the supervision.

\begin{equation}\footnotesize{
\mathcal{L}_{\text{vel}} =
\sum_f
\big\|
w^{(f)} \odot \big((
\hat y_0^{(f)} - \hat y_0^{(f-1)}) - (y^{(f)} - y^{(f-1)})
\big)\big\|_2^2, }
\end{equation}

\noindent where $\odot$ denotes joint-wise multiplication, $w^{(f)} = d^{(f)} \odot \mathbf{1}_{\{d^{(f)} > d_\text{min}\}}$ \amir{(following \cref{sec:method_depth})}, and $\hat y_0 = \Pi_c(\hat x_0)$.

\paragraph{Motion Representation Alignment.}
Motion generation models commonly adopt the motion
representation of \cite{Guo_2022_CVPR}, which includes \amir{root
velocity, joint positions}, joint rotations, joint velocities, and binary foot-contact labels.
This over-parameterized representation often leads to motions that better match human perceptual judgments. Among these, the rotation, joint-velocity, and foot-contact channels are redundant in the sense \amir{that they can be derived directly from the joint positions}. We denote these redundant components collectively by~$\mathbf{r}$ \amir{and the process of deriving them $\Gamma(x)$}. \amir{As MDM operates on the noised concatenation of $x$ and $\mathbf{r}$, we partition its ($A_J+B_J$)-channel output into two components: the first $A_J$ channels represent the predicted root and joint positions $\hat{x}_0$ (used previously), while the remaining $B_J$ channels constitute the redundant representations $\hat{r}_0$ (discussed solely here). Further details in \cref{app:HumanML_chunnels}.}

\begin{figure}[h]
    \centering
    \begin{subfigure}[c]{0.46\linewidth}
        We evaluate \methodname{} in three complementary settings. \cref{sec:synthetic} uses a 2D-only version of HumanML3D, isolating the supervision regime from pose-estimation errors. \cref{sec:PriorExtent} trains on real monocular video from Fit3D, demonstrating that the learned prior extends to fitness motions far outside the lifter's distribution. \cref{sec:nba} compares against MAS on the centered NBA dataset on which MAS was designed to operate, providing a head-to-head comparison under conditions favorable to the baseline.
        
    \end{subfigure}
    \hfill
    \begin{subfigure}[c]{0.5\linewidth}
        \centering
        \includegraphics[width=\linewidth]{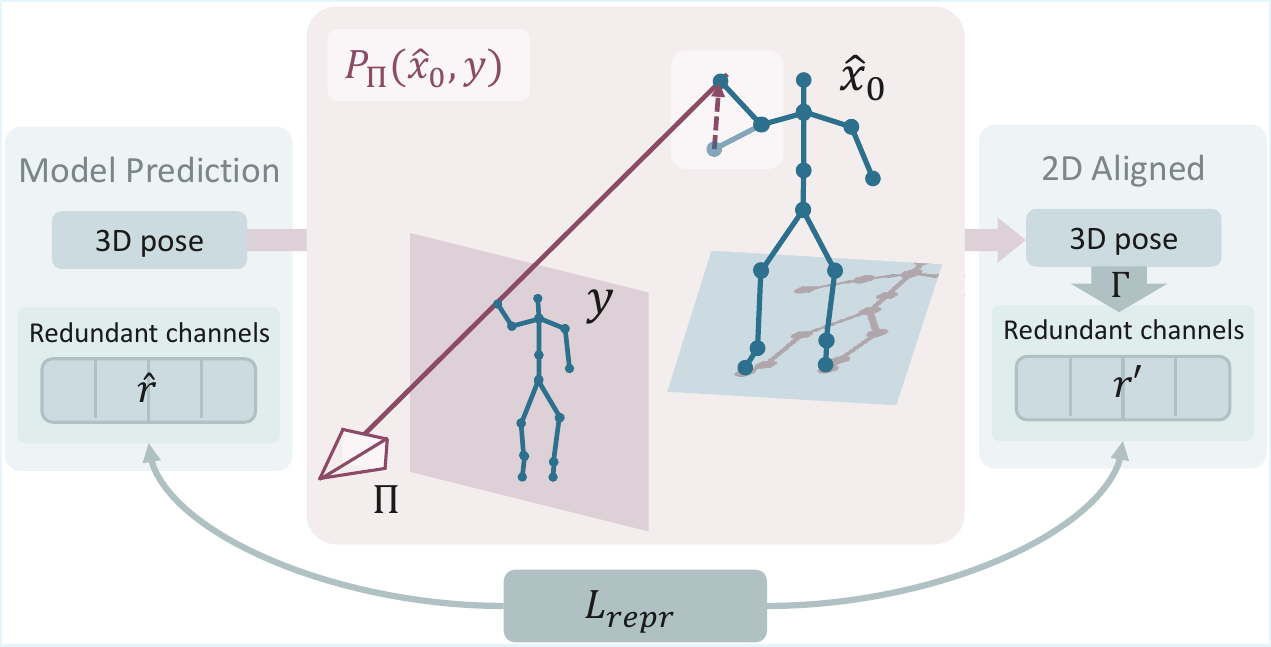}
        \caption{\textbf{Motion Representation Alignment.} An illustration of camera ray projection and its utilization for $\mathcal{L}_{\text{repr}}$. Each joint is projected to the closest point along the ray through its 2D location and the camera center.}
        \label{fig:OverparametrziedLoss}
    \end{subfigure}
\end{figure}

Since no 3D ground-truth supervision is available for $\mathbf{r}$, we derive
2D-consistent pseudo-targets by applying the ray-projection operator \amir{(illustrated in \cref{fig:OverparametrziedLoss} and formally derived in \cref{app:ray_projecting})} to the
predicted 3D motion\amir{, producing 2D aligned 3D motion}, and converting the result to the over-parameterized representation:
\begin{equation}
\mathbf{r}' = \texttt{stop\_gradient}\big(\Gamma(P_\Pi(\hat{x}_0, y))\big),
\end{equation}
where $\Gamma$ calculates the redundant channels from a 3D motion sequence.
We then supervise the corresponding denoised outputs, as shown in
\cref{fig:OverparametrziedLoss}, using
\(
\mathcal{L}_{\text{repr}} = \|\hat{\mathbf{r}}_0 - \mathbf{r}'\|_2^2.
\)
This provides an indirect 2D-based supervisory signal for the redundant
motion channels, helping the model remain consistent with both its own
predictions and the available 2D data throughout generation.

\subsection{Training Objective and Scheme}
\label{sec:method_training}

The overall loss combines our proposed terms:
\begin{equation}
\mathcal{L}_{\text{total}} =
\lambda_{\text{pos}}\mathcal{L}_{\text{pos}} +
\lambda_{\text{vel}}\mathcal{L}_{\text{vel}} +
\lambda_{\text{repr}}\mathcal{L}_{\text{repr}}
\end{equation}

\noindent
Training begins with a warm up phase of pretraining on approximate 3D motions predicted by the lifter $L_\phi$, similar to standard MDM training, to initialize the motion prior. We then adopt the LIS-style schedule with threshold~$t^*$.  
For $t > t^*$, we apply the full loss $\mathcal{L}_{\text{total}}$.  
For $t \le t^*$, we further apply multi-step denoising. This staged procedure stabilizes learning and allows the model to progress from lifter-based supervision to fully 2D-derived constraints.

\pagebreak
\section{Experiments}
\label{sec:experiments}

\paragraph{Implementation Details.} For \cref{sec:synthetic,sec:nba}, we first warm up by pretraining on the lifter's approximate 3D motions for 400K batches of 64 samples to initialize the motion prior, using the lifted samples of the respective dataset, then train with our full objective for an additional 200K batches. For \cref{sec:PriorExtent}, since the dataset is small, we initialize from the model trained on synthetic data (\cref{sec:synthetic}) and fine-tune for 20K batches. We select $\lambda_{\text{vel}}$, $t^*$, \amir{and the number of denoising steps below $t^*$} by Bayesian search over the validation FID on \cref{sec:synthetic} using MVLift~\cite{li2025mvlift} as the lifter; the resulting values (\cref{app:hyperparameters}) are reused \amir{unchanged across all models for the synthetic and Fit3D experiments}. Validation FID stayed within 0.7--2 across most searched configurations, indicating low sensitivity to these choices. \amir{ For the NBA experiment, we perform the same search using a random strategy over the training set FID (as no validation split is available). FID spans 7-11 across all configurations tested.}

\subsection{Text-to-3D Motion from 2D Poses}
\label{sec:synthetic}
HumanML3D~\cite{Guo_2022_CVPR} consists of 14,616 motion sequences sourced from AMASS~\cite{mahmood2019amassarchivemotioncapture} and HumanAct12~\cite{guo2020action2motion}, paired with 44,970 textual descriptions, standardized to 20\,FPS and capped at 10 seconds. We construct a 2D-only version of HumanML3D by sampling random cameras, rendering the 2D motions and lifting them by either MotionBERT~\cite{motionbert2022} or MVLift~\cite{li2025mvlift}. 
For evaluation in 3D we follow T2M \cite{Guo_2022_CVPR}, using MDM \cite{tevet2023human} inference and evaluation code.  
Further details in \cref{app:exp_tech_details}.

\paragraph{Compared Methods.} We compare three classes of methods, distinguished by the supervision they require. \emph{3D-supervised upper bound:} MDM~\cite{tevet2023human} trained on the original 3D motions. \emph{2D-supervised baselines:} MAS~\cite{kapon2024mas}, trained as a text-conditioned, non-centered variant; and MDM trained on lifter outputs (MotionBERT or MVLift) treated as 3D ground truth --- these baselines use no camera information. \emph{Our method:} three variants that supervise in 2D via reprojection and therefore require camera parameters. \emph{Ours/MotionBERT} and \emph{Ours/MVLift} use ground-truth cameras with their corresponding lifters as teachers. \emph{Ours/MVLift (PnP)} estimates the camera via PnP from the MVLift teacher (see \cref{PNPdetails}), matching the information available to the 2D-supervised baselines.


\paragraph{Results.} As shown in \cref{tab:synthetic}, under matched supervision \emph{Ours/MVLift\,(PnP)} reduces FID by 0.21 against \emph{MDM/MVLift}, demonstrating the value of 2D-reprojection training over training MDM directly on lifted 3D. With GT cameras the advantage grows: \emph{Ours/MVLift} reaches FID 0.88, nearly closing the gap to the 3D-supervised upper bound (0.54), empirically supporting the loss-equivalence claim of \cref{sec:method_depth}, highlighting that the remaining gap is attributed to the camera estimation error. With improved camera estimation methods, our method directly benefits. 

\begin{figure*}[h]
    \centering
    \includegraphics[width=1\linewidth]{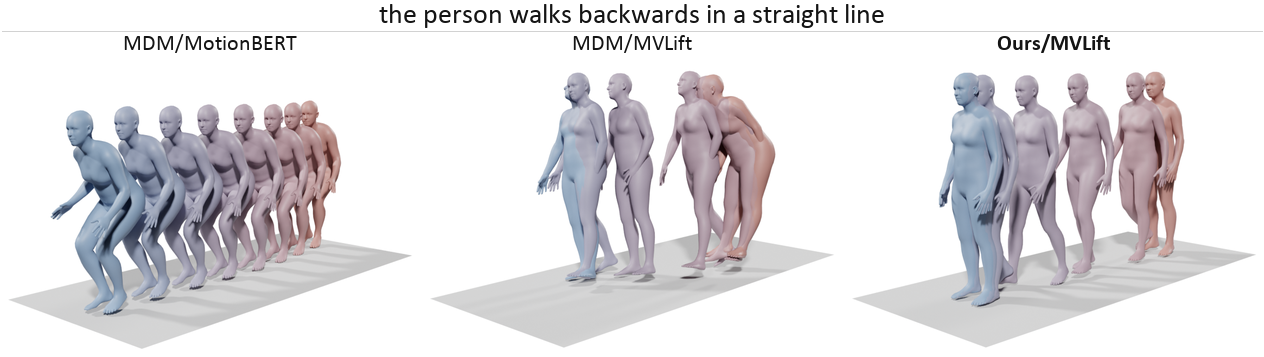}
    \caption{Qualitative comparison on HumanML3D for the prompt ``the person walks backwards in a straight line''. Frames progress from blue to red. The MDM-\gal{trained-}on-lifter baselines inherit teacher artifacts (foot sliding, drift\amir{ing, unrealistic poses}), while \emph{VideoMDM} generates a clean trajectory consistent with the prompt.}
    \label{fig:text23d_humanml}
\end{figure*}

\cref{fig:text23d_humanml} confirms the quantitative picture: \emph{MDM/MotionBERT} inherits lifter artifacts and exhibits sliding, \emph{MDM/MVLift} is better but still lacks coherence, while \emph{VideoMDM} produces clean trajectories that accurately follow the prompt. Additional results are available in the supplementary material.



\newcommand{\smallbox}[2]{{%
  \setlength{\fboxsep}{1pt}
  \colorbox{#1}{\raisebox{0pt}[1.0ex][0.0ex]{#2}}
}}

\begin{table*}[t]
    \centering
    \begin{tabular*}{\textwidth}{@{\extracolsep{\fill}}lccccc}
        \toprule
        \textbf{Method} & \textbf{FID}$\downarrow$ & \textbf{Diversity}$\rightarrow$ & \textbf{R Precision} & \textbf{Multimodal} & \textbf{Multimodality}$\uparrow$ \\
        & & & \textbf{(top 3)}$\uparrow$ & \textbf{Dist}$\downarrow$ & \\
        \midrule
        Ground-truth & $.0016^{\pm .000}$ & $9.459^{\pm .052}$ & $0.796^{\pm .002}$ & $2.975^{\pm .009}$ & --- \\
        MDM (3D data) & $0.544^{\pm .044}$ & $9.559^{\pm .086}$ & $0.611^{\pm .007}$ & $5.566^{\pm .027}$ & $2.799^{\pm .072}$ \\
        \midrule
        MAS & $22.056^{\pm .009}$ & $6.236^{\pm .051}$ & $0.383^{\pm .000}$ & $6.416^{\pm .000}$ & --- \\
        MDM/MotionBERT & $5.660^{\pm .156}$ & $8.198^{\pm .048}$ & $0.666^{\pm .006}$ & $4.112^{\pm .016}$ & \cellcolor{tabthird}$2.453^{\pm .110}$ \\
        MDM/MVLift & $1.671^{\pm .094}$ & $8.793^{\pm .048}$ & \cellcolor{tabsecond}$0.719^{\pm .005}$ & \cellcolor{tabsecond}$3.514^{\pm .018}$ & $2.375^{\pm .105}$ \\
        Ours/MVLift (PnP) & \cellcolor{tabthird}$1.462^{\pm .097}$ & \cellcolor{tabthird}$9.130^{\pm .052}$ & \cellcolor{tabthird}$0.714^{\pm .007}$ & \cellcolor{tabthird}$3.527^{\pm .031}$ & \cellcolor{tabfirst}$2.692^{\pm .087}$ \\
        Ours/MotionBERT & \cellcolor{tabsecond}$1.454^{\pm .090}$ & \cellcolor{tabfirst}$9.533^{\pm .060}$ & $0.681^{\pm .007}$ & $3.677^{\pm .030}$ & \cellcolor{tabsecond}$2.457^{\pm .050}$ \\
        Ours/MVLift & \cellcolor{tabfirst}$0.876^{\pm .090}$ & \cellcolor{tabsecond}$9.630^{\pm .068}$ & \cellcolor{tabfirst}$0.721^{\pm .005}$ & \cellcolor{tabfirst}$3.450^{\pm .028}$ & $2.449^{\pm .110}$ \\
        \bottomrule
    \end{tabular*}
    \caption{\label{tab:synthetic} Text-to-motion models trained in 2D, evaluated on the HumanML3D test split. \smallbox{tabfirst}{\strut Red}, \smallbox{tabsecond}{\strut orange}, and \smallbox{tabthird}{\strut yellow} indicate 1st, 2nd, and 3rd place per column among 2D-supervised methods. \emph{X/Y} denotes a method \emph{X} trained on lifter \emph{Y}. \emph{Ours} variants use ground-truth cameras unless marked (PnP). Remarkably, \emph{Ours/MVLift} achieves performance only 0.332 FID away from 3D-supervised MDM.}
\end{table*}

\subsection{Real-Video Training: Beyond the Lifter Distribution}
\label{sec:PriorExtent}

Our central claim is that 2D supervision from monocular video unlocks 3D motion distributions inaccessible to MoCap-bound training. We test this on Fit3D~\cite{Fieraru_2021_CVPR}: real videos using 2D-pose extraction, and motions far outside the distribution captured by the lifters. Fit3D contains 611 training sequences across 37 fitness exercises, with synchronized video and accurate 3D ground truth used only for evaluation. Many of these motions have no analog in HumanML3D (e.g. mule kicks, burpees, stretches). For each sequence we randomly select one camera, extract 2D poses with RTMPose~\cite{jiang2023rtmposerealtimemultipersonpose}. Approximate 3D poses are obtained from the same video using WHAM~\cite{wham:cvpr:2024}, a dedicated video-to-3D method, serving as the noisy teacher. Camera positions are provided in the dataset or estimated using PnP. 
We initialize \methodname{} from Ours/MVLift trained in \cref{sec:synthetic} and fine-tune for 20K batches. Further details in \cref{app:exp_tech_details}.

\paragraph{Evaluation.} 
Fit3D is too small for reliable generative metrics (FID). We instead evaluate \methodname{} in two ways: as a 2D-to-3D lifter~\cref{tab:lifting}, and with a human survey~\cref{tab:human_survey_fit3d}. We repurpose \methodname{} as a lifter via inference-time guidance: at every denoising step, we project the predicted clean motion onto the camera rays of the observed 2D keypoints, $\hat{x}_0 \leftarrow P_{\Pi}(\hat{x}_0, y)$ (\cref{app:formal_guidance}); this provides a direct probe of prior quality against corresponding ground-truth 3D motions. 

\paragraph{Metrics.} We report Mean Per-Joint Position Error (MPJPE) in mm, Procrustes-aligned MPJPE (PA-MPJPE) in mm using per-frame scale, rotation, and translation alignment, Percentage of Correct Keypoints (PCK) at 50\,mm and 100\,mm thresholds, and acceleration error (m/s$^2$) from second-order finite differences. For generative quality we report KID~\cite{binkowski2018KID} -- which is considered more reliable for smaller sample size -- in the HumanML3D evaluation VAE~\cite{Guo_2022_CVPR} , using subsampling with replacement (100 trials) for stable means and standard deviations.


\paragraph{Compared Methods.} On 2D-to-3D motion lifting, we compare against \emph{WHAM}~\cite{wham:cvpr:2024},  \emph{MVLift}~\cite{li2025mvlift}, and MDM trained on HumanML3D guided in the same way as ours. For human preference we compare against  generative text-to-3D methods: MDM/WHAM and MDM/MVLift and on the lifted motions themselves. 
%


\paragraph{Results -- motion lifting.} As shown in \cref{tab:lifting}, \emph{Ours} achieves the best results across most metrics: MPJPE drops by a factor of 2 and 4 compared to WHAM and MDM, respectively. \emph{Ours\,(PnP)} retains most of this gap. Crucially, both our variants excel in Accel and KID, indicating smooth motions that are statistically aligned with the true 3D motion distribution. WHAM retains an edge on PA-MPJPE, which removes scale and rotation and rewards local pose accuracy --- consistent with WHAM's pose-supervised objective and its use of image features, which carry local detail unavailable in the 2D skeleton alone.

\begin{figure}[h]
    \centering
    \begin{subfigure}[c]{0.52\linewidth}
        \paragraph{Results -- text-to-motion.} For reliable evaluation of conditional 3D motion in the absence of large scale data, we turn to human inspection. \cref{tab:human_survey_fit3d} reports the percentage of human preferred motions between each baseline and ours (e.g. 40.0\% on MDM/WHAM means ours was preferred 60.0\% of the time). While Ours/WHAM consistently outperforms all baselines, Ours/WHAM (PnP) loses to the WHAM baselines. Inspecting the down-voted motions, we observed these suffered from poor alignment with the ground, which was visually unsatisfying. We hypothesize that a ground-aware inference or post-processing could significantly boost visual appearance, yet we left the raw results for full transparency.  
    \end{subfigure}
    \hfill
    \begin{subfigure}[c]{0.46\linewidth}
        \centering
        \setlength{\tabcolsep}{3pt}
        \begin{tabular}{lcc}
        \toprule
        \textbf{Method} & \textbf{Ours/WHAM} & \textbf{Ours/WHAM} \\
        \textbf{} & \textbf{pref.} & \textbf{(PNP) pref.} \\
        \midrule
        MDM/WHAM       & 40.0\% & 62.1\% \\
        WHAM           & 38.7\% & 60.3\% \\
        MVLift         & 25.0\% & 42.3\% \\
        MDM/MVLift     & 11.3\% & 12.5\% \\
        \bottomrule
        \end{tabular}
        \caption{\amir{Human Preference Survey results for the Fit3D test set text prompts. MDM-based models (Ours, MDM/WHAM, and MDM/MVLift) receive only text prompts; WHAM and MVLift also receive the original video.}\label{tab:human_survey_fit3d}}
    \end{subfigure}
\end{figure}

\cref{fig:fit3dGen} shows an example of a Fit3D exercise prompt. Ours/WHAM produces a coherent and text-aligned motion, while baselines suffer from misalignment to this motion which is outside their train set. Additional video examples are in the supplementary material.

\begin{table*}[t]
\centering
\setlength{\tabcolsep}{2pt}
\begin{tabular}{@{}lcccccc@{}}
\hline
Model & MPJPE & PA-MPJPE & PCK@ & PCK@ & Accel & KID \\
& (mm) & (mm) & 50mm(\%) & 100mm(\%) & (m/s$^2$) & ($\pm$ std) \\
\hline
MDM        & 440.33          & 114.06          & 6.34           & 24.20          & 7.71  & $0.050 \pm 0.010$ \\
WHAM\;     & 228.47          & \textbf{51.12}  & 2.89           & 17.39          & 17.66 & $0.063 \pm 0.008$ \\
MVLift\;   & 283.06          & 94.45           & 5.84           & 22.99          & 3.14  & $0.028 \pm 0.006$ \\
Ours/WHAM (PnP) & 185.81          & 74.03           & 15.49          & 52.82          & \textbf{3.04}  & $0.013 \pm 0.003$ \\
Ours/WHAM       & \textbf{111.24} & 61.69           & \textbf{22.26} & \textbf{62.80} & 3.16  & $\mathbf{0.011 \pm 0.003}$ \\
\hline
\end{tabular}
\caption{Lifting evaluation on Fit3D held out subject. Our method achieves best KID and MPJPE both on GT and PnP cameras.}
\label{tab:lifting}
\end{table*}

\begin{figure}[h]
    \centering
    \begin{subfigure}[c]{0.54\linewidth}
        \centering
        \includegraphics[width=\linewidth]{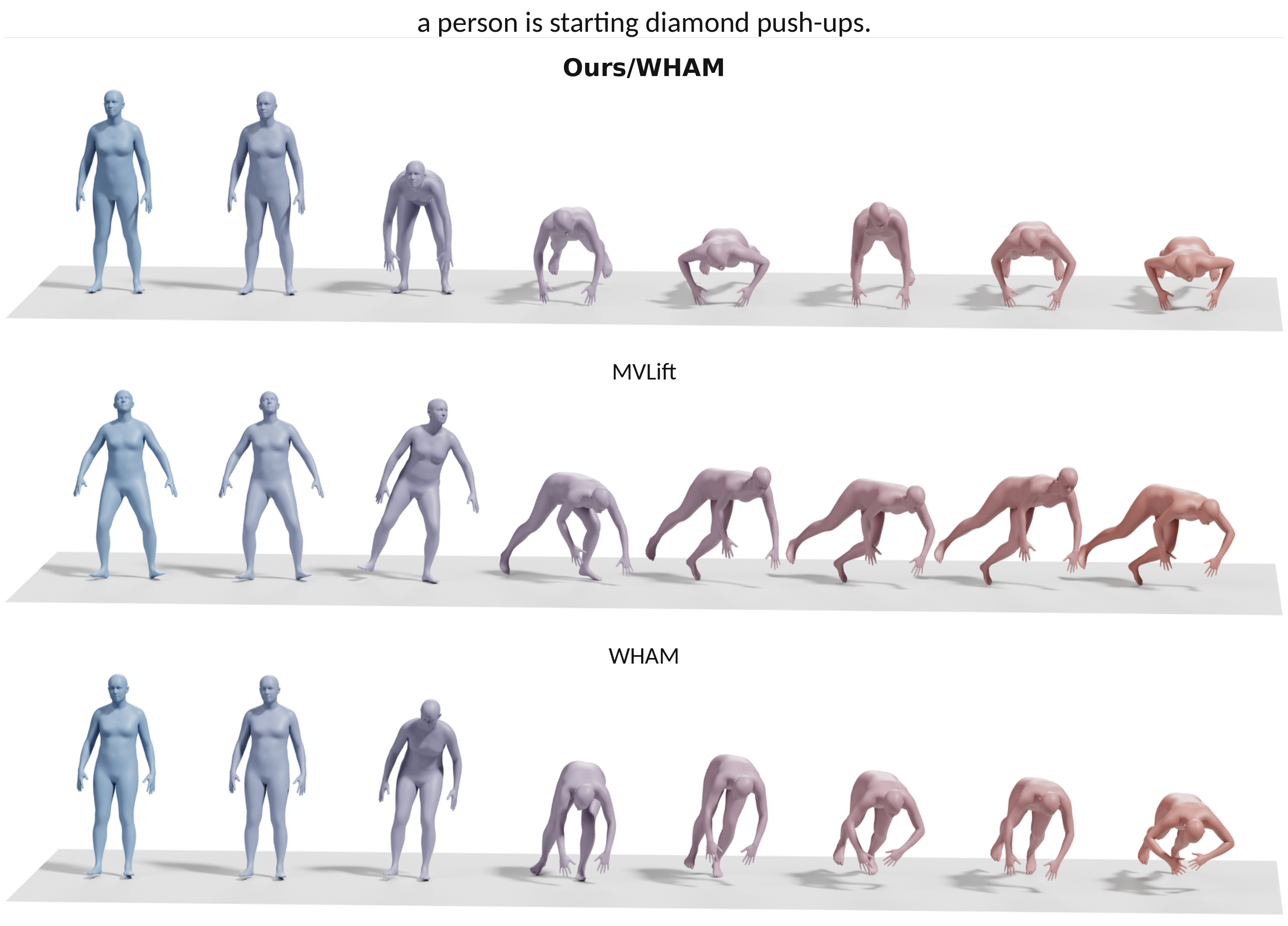}
        \caption{Results of generation by VideoMDM trained on Fit3D \amir{against the lifter baselines, the lifters having access to the 2D data while Ours only to the text prompt.}}
        \label{fig:fit3dGen}
    \end{subfigure}
    \hfill
    \begin{subfigure}[c]{0.43\linewidth}
        \centering
        \small
        \setlength{\tabcolsep}{3pt}
        \renewcommand\theadfont{\normalfont\small}
        \begin{tabular}{lcccc}
        \toprule
        \thead[l]{w/o\\Variant} & \thead{FID\\$\downarrow$} & \thead{Diversity\\$\rightarrow$} & \thead{R-Prec\\$\uparrow$} & \thead{MM-Dist\\$\downarrow$} \\
        \midrule
        \makecell[l]{distance\\weighting}             & 1.27 & 8.84 & \textbf{0.72} & \textbf{3.48} \\
        \makecell[l]{multistep\\for $t < t^*$}        & 9.85 & 7.15 & 0.45          & 5.60 \\
        $\mathcal{L}_{\text{vel}}$                     & 1.58 & 9.19 & 0.69          & 3.62 \\
        $\mathcal{L}_{\text{repr}}$                    & 5.75 & 8.25 & 0.60          & 4.42 \\
        \makecell[l]{ray proj \\ for $\mathcal{L}_{\text{repr}}$} & 2.72 & 8.91 & 0.63 & 4.09 \\
        Ours                                           & \textbf{1.05} & \textbf{9.60} & 0.71          & 3.50 \\
        \bottomrule
        \end{tabular}
        \caption{Ablations of our method using PnP cameras and MVLift as a lifter, mean values over 5 replications on the HumanML3D validation set.}
        \label{tab:ablations}
    \end{subfigure}
\end{figure}

\subsection{Unconditional 3D Generation from NBA videos}
\label{sec:nba}

The NBA dataset, released with MAS~\cite{kapon2024mas}, provides single-view basketball sequences with 16-joint AlphaPose~\cite{10.1109/TPAMI.2022.3222784} detections. Since MAS was designed for centered motion generation, the dataset was collected accordingly: all motions are centered per frame (root at the origin), and no text prompts are provided. The release also includes lifted centered motions from MotionBERT~\cite{motionbert2022} and ElePose~\cite{elepose2022cvpr}. We use the latter as our noisy teacher. Further technical details are in \cref{app:exp_tech_details}.



\paragraph{Methods and Results.} We compare against MAS and report results in \cref{tab:nba}. The standard evaluation protocol, released with MAS, computes metrics in the embedding of a 2D VAE trained on the same data. Inspecting the VAE's encode--decode reconstructions reveals lower fidelity than the corresponding HumanML3D evaluation VAE, suggesting the VAE-based metrics here are a less reliable indicator of perceptual quality. We therefore conducted a human preference study (\cref{app:human_survey}) in which participants compared our VideoMDM and MAS outputs in pairwise settings. Across 200 votes, VideoMDM was preferred in nearly two thirds, aligning with our own visual inspection of the generations. Generated motion examples for both models are included in the supplementary material.

For completeness, we also report the standard MAS protocol metrics. We further note an issue with how Recall is computed in this protocol: for each real sample, it asks whether some generated sample lies within a threshold defined by the \emph{generated}-distribution spread, biasing the metric toward exaggerated diversity. We additionally report Recall$^\dagger$ (\cref{app:nba_extra}), which anchors the threshold to the real-distribution spread instead and better reflects whether the generated motions cover the real distribution. VideoMDM falls slightly behind MAS on FID and Diversity, but achieves substantial gains in Precision and the best Recall$^\dagger$, indicating that our generations spread less diversely, but otherwise similarly to the real distribution in the VAE embedding.


\begin{table*}[t]
\centering
\begin{tabular}{lcccccc}
\toprule
\textbf{Method} & \textbf{Human} $\uparrow$ & \textbf{FID}$\downarrow$ & \textbf{Diversity}$\rightarrow$ & \textbf{Precision}$\uparrow$ & \textbf{Recall}$\uparrow$ & \textbf{Recall}$^\dagger\uparrow$ \\
 & \textbf{Pref.} & & & & & \\
\midrule
Training Data        & --- & $1.05^{\pm .02}$ & $8.97^{\pm .05}$ & $0.73^{\pm .01}$ & $0.73^{\pm .01}$ & $0.86^{\pm .01}$ \\
ElePose & --- & $10.76^{\pm .45}$ & $9.72^{\pm .05}$ & $0.28^{\pm .02}$ & $0.58^{\pm .03}$ & $0.45^{\pm .01}$ \\
MotionBERT & --- & $30.22^{\pm .26}$ & $9.57^{\pm .09}$ & $0.04^{\pm .00}$ & $0.34^{\pm .04}$ & $0.04^{\pm .01}$ \\
MAS & 36.0\% & $\mathbf{5.38^{\pm .06}}$ & $\mathbf{9.47^{\pm .06}}$ & $0.50^{\pm .01}$ & $\mathbf{0.60^{\pm .01}}$ & $0.68^{\pm .00}$ \\
Ours/ElePose & $\mathbf{64.0\%}$ & $7.18^{\pm .09}$ & $7.93^{\pm .04}$ & $\mathbf{0.94^{\pm .00}}$ & $0.10^{\pm .00}$ & $\mathbf{0.89^{\pm .01}}$ \\
\bottomrule
\end{tabular}
\caption{Evaluation on the NBA dataset. Human preference is reported between MAS and our VideoMDM. Other metrics follow MAS~\cite{kapon2024mas}. VideoMDM is preferred nearly two thirds of votes.}
\label{tab:nba}
\end{table*}


\vspace{-7pt}
\subsection{Ablations}
\vspace{-7pt}

We ablate each component of \methodname{} on the HumanML3D validation set, using PnP-estimated cameras with MVLift as the teacher. Each row in \cref{tab:ablations} reports the mean over 5 replicate runs with one component removed: \emph{distance weighting} drops the $\mathcal{L}_{\text{pos}}$ loss defined in (\cref{sec:method_depth}); \emph{multistep for $t < t^*$} disables multistep denoising entirely (i.e. $t^* = 0$) and trains using the noised lifter prediction directly across all $t$~\cite{peng2025lesson}; \emph{$\mathcal{L}_{\text{vel}}$} and \emph{$\mathcal{L}_{\text{repr}}$} drop the corresponding losses by setting $\lambda_{\text{vel}} = 0$ or $\lambda_{\text{repr}} = 0$; and \emph{ray proj} replaces the camera-ray projection used in $\mathcal{L}_{\text{repr}}$ with a direct comparison $\mathcal{L}_{\text{repr}} = \|\hat{r} - \Gamma(\hat{x}_0)\|$. Multistep denoising and $\mathcal{L}_{\text{repr}}$ are the two most impactful components. Replacing the ray projection in $\mathcal{L}_{\text{repr}}$ with a naive 3D comparison roughly triples FID, confirming that our 2D-consistent pseudo-targets are essential to the representation alignment. $\mathcal{L}_{\text{vel}}$ and distance weighting contribute smaller but non-negligible improvements on FID, with negligible differences on the remaining metrics.

\vspace{-8pt}
\section{Conclusion, Limitations and Future Work}
\label{sec:limitation}
 \vspace{-8pt}


We have presented VideoMDM, a training method for 3D human motion diffusion using only 2D supervision, and demonstrated its ability to generate high-quality motions that in some settings nearly match the performance of fully 3D-supervised methods. Our cross-modality diffusion owes its success to a set of stabilization techniques and natural motion regularizations formulated in 2D. VideoMDM makes a significant stride towards training motion diffusion models directly from abundant monocular videos. Such capability opens up new possibilities for learning generative priors of real-world motion distributions such as multi-person behaviors and human-object interactions which are otherwise difficult to acquire.


\paragraph{Limitations.} VideoMDM achieves its strongest results with ground-truth camera parameters, which are unavailable for most in-the-wild videos. PnP-estimated cameras recover most of the gap on synthetic HumanML3D but leave a larger drop on Fit3D, where camera estimation noise compounds with lifter and pose-extraction noise; better camera estimators would directly translate into better priors. Our method also depends on a pretrained 2D-to-3D lifter as a noisy teacher: although the learned prior generalizes substantially beyond the lifter's distribution (\cref{sec:PriorExtent}), domains where no reasonable lifter is available (e.g.\ non-human motion) remain out of reach. All settings we evaluate contain no or only minimal occlusions; extending to occluded settings such as those in~\cite{grauman2024egoexo4dunderstandingskilledhuman} is necessary for fully in-the-wild deployment.


\begin{ack}
Or Litany acknowledges support from the Israel Science Foundation (grant 624/25) and the Azrieli Foundation Early Career Faculty Fellowship. This research was also supported in part by an academic gift from Meta. The authors gratefully acknowledge this support. This research was supported by the Council for Higher Education in Israel under the Moonshot Project.
\end{ack}

\bibliography{main}

@String(CVPR= {IEEE Conf. Comput. Vis. Pattern Recog.})

@String(ICCV= {Int. Conf. Comput. Vis.})

@String(ECCV= {Eur. Conf. Comput. Vis.})

@String(NIPS= {Adv. Neural Inform. Process. Syst.})

@String(CVPR  = {CVPR})

@String(ICCV  = {ICCV})

@String(ECCV  = {ECCV})

@String(NIPS  = {NeurIPS})

@inproceedings{peng2025lesson,
  title={A Lesson in Splats: Teacher-Guided Diffusion for 3D Gaussian Splats Generation with 2D Supervision},
  author={Peng, Chensheng and Sobol, Ido and Tomizuka, Masayoshi and Keutzer, Kurt and Xu, Chenfeng and Litany, Or},
  booktitle={Proceedings of the IEEE/CVF International Conference on Computer Vision},
  year={2025}
}

@inproceedings{kapon2024mas,
  title={Mas: Multi-view ancestral sampling for 3d motion generation using 2d diffusion},
  author={Kapon, Roy and Tevet, Guy and Cohen-Or, Daniel and Bermano, Amit H},
  booktitle={Proceedings of the IEEE/CVF Conference on Computer Vision and Pattern Recognition},
  pages={1965--1974},
  year={2024}
}

@InProceedings{Guo_2025_ICCV,
    author    = {Guo, Ruoxi and Pi, Huaijin and Shen, Zehong and Shuai, Qing and Hu, Zechen and Wang, Zhumei and Dong, Yajiao and Hu, Ruizhen and Komura, Taku and Peng, Sida and Zhou, Xiaowei},
    title     = {Motion-2-to-3: Leveraging 2D Motion Data for 3D Motion Generations},
    booktitle = {Proceedings of the IEEE/CVF International Conference on Computer Vision (ICCV)},
    month     = {October},
    year      = {2025},
    pages     = {14305-14316}
}

@inproceedings{
tevet2023human,
title={Human Motion Diffusion Model},
author={Guy Tevet and Sigal Raab and Brian Gordon and Yoni Shafir and Daniel Cohen-or and Amit Haim Bermano},
booktitle={The Eleventh International Conference on Learning Representations },
year={2023},
url={https://openreview.net/forum?id=SJ1kSyO2jwu}
}

@inproceedings{wham:cvpr:2024, 
  title = {{WHAM}: Reconstructing World-grounded Humans with Accurate {3D} Motion},
  author = {Shin, Soyong and Kim, Juyong and Halilaj, Eni and Black, Michael J.}, 
  booktitle = {IEEE/CVF Conf.~on Computer Vision and Pattern Recognition (CVPR)},
  month = jun,
  year = {2024},
  doi = {},
  month_numeric = {6}, 
}

@inproceedings{li2024coin,
    title={COIN: Control-Inpainting Diffusion Prior for Human and Camera Motion Estimation},
    author={Li, Jiefeng and Yuan, Ye and Rempe, Davis and Zhang, Haotian and Lu, Cewu and Kautz, Jan and Iqbal, Umar},
    booktitle={European Conference on Computer Vision (ECCV)},
    year={2024}
}

@inproceedings{wang2024tram,
  title={TRAM: Global Trajectory and Motion of 3D Humans from in-the-wild Videos},
  author={Wang, Yufu and Wang, Ziyun and Liu, Lingjie and Daniilidis, Kostas},
  booktitle={European Conference on Computer Vision},
  pages={467--487},
  year={2024},
  organization={Springer}
}

@misc{zhang2024rohmrobusthumanmotion,
      title={RoHM: Robust Human Motion Reconstruction via Diffusion}, 
      author={Siwei Zhang and Bharat Lal Bhatnagar and Yuanlu Xu and Alexander Winkler and Petr Kadlecek and Siyu Tang and Federica Bogo},
      year={2024},
      eprint={2401.08570},
      archivePrefix={arXiv},
      primaryClass={cs.CV},
      url={https://arxiv.org/abs/2401.08570}, 
}

@inproceedings{motionbert2022,
  title     =   {MotionBERT: A Unified Perspective on Learning Human Motion Representations}, 
  author    =   {Zhu, Wentao and Ma, Xiaoxuan and Liu, Zhaoyang and Liu, Libin and Wu, Wayne and Wang, Yizhou},
  booktitle =   {Proceedings of the IEEE/CVF International Conference on Computer Vision},
  year      =   {2023},
}

@INPROCEEDINGS {elepose2022cvpr,
author = { Wandt, Bastian and Little, James J. and Rhodin, Helge },
booktitle = { 2022 IEEE/CVF Conference on Computer Vision and Pattern Recognition (CVPR) },
title = {{ ElePose: Unsupervised 3D Human Pose Estimation by Predicting Camera Elevation and Learning Normalizing Flows on 2D Poses }},
year = {2022},
pages = {6625-6635},
doi = {10.1109/CVPR52688.2022.00652},
url = {https://doi.ieeecomputersociety.org/10.1109/CVPR52688.2022.00652},
publisher = {IEEE Computer Society},
address = {Los Alamitos, CA, USA},
month =Jun
}

@inproceedings{li2025mvlift,
  title={Lifting motion to the 3d world via 2d diffusion},
  author={Li, Jiaman and Liu, C Karen and Wu, Jiajun},
  booktitle={Proceedings of the Computer Vision and Pattern Recognition Conference},
  pages={17518--17528},
  year={2025}
}

@misc{ramesh2022hierarchicaltextconditionalimagegeneration,
      title={Hierarchical Text-Conditional Image Generation with CLIP Latents}, 
      author={Aditya Ramesh and Prafulla Dhariwal and Alex Nichol and Casey Chu and Mark Chen},
      year={2022},
      eprint={2204.06125},
      archivePrefix={arXiv},
      primaryClass={cs.CV},
      url={https://arxiv.org/abs/2204.06125}, 
}

@misc{ho2020denoisingdiffusionprobabilisticmodels,
      title={Denoising Diffusion Probabilistic Models}, 
      author={Jonathan Ho and Ajay Jain and Pieter Abbeel},
      year={2020},
      eprint={2006.11239},
      archivePrefix={arXiv},
      primaryClass={cs.LG},
      url={https://arxiv.org/abs/2006.11239}, 
}

@misc{zhang2022motiondiffusetextdrivenhumanmotion,
      title={MotionDiffuse: Text-Driven Human Motion Generation with Diffusion Model}, 
      author={Mingyuan Zhang and Zhongang Cai and Liang Pan and Fangzhou Hong and Xinying Guo and Lei Yang and Ziwei Liu},
      year={2022},
      eprint={2208.15001},
      archivePrefix={arXiv},
      primaryClass={cs.CV},
      url={https://arxiv.org/abs/2208.15001}, 
}

@inproceedings{chen2023executing,
  title={Executing your Commands via Motion Diffusion in Latent Space},
  author={Chen, Xin and Jiang, Biao and Liu, Wen and Huang, Zilong and Fu, Bin and Chen, Tao and Yu, Gang},
  booktitle={Proceedings of the IEEE/CVF Conference on Computer Vision and Pattern Recognition},
  pages={18000--18010},
  year={2023}
}

@misc{zou2024parcopartcoordinatingtexttomotionsynthesis,
      title={ParCo: Part-Coordinating Text-to-Motion Synthesis}, 
      author={Qiran Zou and Shangyuan Yuan and Shian Du and Yu Wang and Chang Liu and Yi Xu and Jie Chen and Xiangyang Ji},
      year={2024},
      eprint={2403.18512},
      archivePrefix={arXiv},
      primaryClass={cs.CV},
      url={https://arxiv.org/abs/2403.18512}, 
}

@misc{karunratanakul2023guidedmotiondiffusioncontrollable,
      title={Guided Motion Diffusion for Controllable Human Motion Synthesis}, 
      author={Korrawe Karunratanakul and Konpat Preechakul and Supasorn Suwajanakorn and Siyu Tang},
      year={2023},
      eprint={2305.12577},
      archivePrefix={arXiv},
      primaryClass={cs.CV},
      url={https://arxiv.org/abs/2305.12577}, 
}

@misc{guo2023momaskgenerativemaskedmodeling,
      title={MoMask: Generative Masked Modeling of 3D Human Motions}, 
      author={Chuan Guo and Yuxuan Mu and Muhammad Gohar Javed and Sen Wang and Li Cheng},
      year={2023},
      eprint={2312.00063},
      archivePrefix={arXiv},
      primaryClass={cs.CV},
      url={https://arxiv.org/abs/2312.00063}, 
}

@misc{pinyoanuntapong2024bammbidirectionalautoregressivemotion,
      title={BAMM: Bidirectional Autoregressive Motion Model}, 
      author={Ekkasit Pinyoanuntapong and Muhammad Usama Saleem and Pu Wang and Minwoo Lee and Srijan Das and Chen Chen},
      year={2024},
      eprint={2403.19435},
      archivePrefix={arXiv},
      primaryClass={cs.CV},
      url={https://arxiv.org/abs/2403.19435}, 
}

@misc{hong2025bipobidirectionalpartialocclusion,
      title={BiPO: Bidirectional Partial Occlusion Network for Text-to-Motion Synthesis}, 
      author={Seong-Eun Hong and Soobin Lim and Juyeong Hwang and Minwook Chang and Hyeongyeop Kang},
      year={2025},
      eprint={2412.00112},
      archivePrefix={arXiv},
      primaryClass={cs.CV},
      url={https://arxiv.org/abs/2412.00112}, 
}

@InProceedings{Guo_2022_CVPR,
    author    = {Guo, Chuan and Zou, Shihao and Zuo, Xinxin and Wang, Sen and Ji, Wei and Li, Xingyu and Cheng, Li},
    title     = {Generating Diverse and Natural 3D Human Motions From Text},
    booktitle = {Proceedings of the IEEE/CVF Conference on Computer Vision and Pattern Recognition (CVPR)},
    year      = {2022},
    pages     = {5152-5161}
}

@misc{bie2022hitdvaehumanmotiongeneration,
      title={HiT-DVAE: Human Motion Generation via Hierarchical Transformer Dynamical VAE}, 
      author={Xiaoyu Bie and Wen Guo and Simon Leglaive and Lauren Girin and Francesc Moreno-Noguer and Xavier Alameda-Pineda},
      year={2022},
      eprint={2204.01565},
      archivePrefix={arXiv},
      primaryClass={cs.CV},
      url={https://arxiv.org/abs/2204.01565}, 
}

@misc{mahmood2019amassarchivemotioncapture,
      title={AMASS: Archive of Motion Capture as Surface Shapes}, 
      author={Naureen Mahmood and Nima Ghorbani and Nikolaus F. Troje and Gerard Pons-Moll and Michael J. Black},
      year={2019},
      eprint={1904.03278},
      archivePrefix={arXiv},
      primaryClass={cs.CV},
      url={https://arxiv.org/abs/1904.03278}, 
}

@inproceedings{guo2020action2motion,
  title={Action2motion: Conditioned generation of 3d human motions},
  author={Guo, Chuan and Zuo, Xinxin and Wang, Sen and Zou, Shihao and Sun, Qingyao and Deng, Annan and Gong, Minglun and Cheng, Li},
  booktitle={Proceedings of the 28th ACM International Conference on Multimedia},
  pages={2021--2029},
  year={2020}
}

@InProceedings{Fieraru_2021_CVPR,
    author = {Fieraru, Mihai and Zanfir, Mihai and Pirlea, Silviu-Cristian and Olaru, Vlad and Sminchisescu, Cristian},
    title = {AIFit: Automatic 3D Human-Interpretable Feedback Models for Fitness Training},
    booktitle = {The IEEE/CVF Conference on Computer Vision and Pattern Recognition (CVPR)},
    month = {June},
    year = {2021}
}

@misc{grauman2024egoexo4dunderstandingskilledhuman,
      title={Ego-Exo4D: Understanding Skilled Human Activity from First- and Third-Person Perspectives}, 
      author={Kristen Grauman and Andrew Westbury and Lorenzo Torresani and Kris Kitani and Jitendra Malik and Triantafyllos Afouras and Kumar Ashutosh and Vijay Baiyya and Siddhant Bansal and Bikram Boote and Eugene Byrne and Zach Chavis and Joya Chen and Feng Cheng and Fu-Jen Chu and Sean Crane and Avijit Dasgupta and Jing Dong and Maria Escobar and Cristhian Forigua and Abrham Gebreselasie and Sanjay Haresh and Jing Huang and Md Mohaiminul Islam and Suyog Jain and Rawal Khirodkar and Devansh Kukreja and Kevin J Liang and Jia-Wei Liu and Sagnik Majumder and Yongsen Mao and Miguel Martin and Effrosyni Mavroudi and Tushar Nagarajan and Francesco Ragusa and Santhosh Kumar Ramakrishnan and Luigi Seminara and Arjun Somayazulu and Yale Song and Shan Su and Zihui Xue and Edward Zhang and Jinxu Zhang and Angela Castillo and Changan Chen and Xinzhu Fu and Ryosuke Furuta and Cristina Gonzalez and Prince Gupta and Jiabo Hu and Yifei Huang and Yiming Huang and Weslie Khoo and Anush Kumar and Robert Kuo and Sach Lakhavani and Miao Liu and Mi Luo and Zhengyi Luo and Brighid Meredith and Austin Miller and Oluwatumininu Oguntola and Xiaqing Pan and Penny Peng and Shraman Pramanick and Merey Ramazanova and Fiona Ryan and Wei Shan and Kiran Somasundaram and Chenan Song and Audrey Southerland and Masatoshi Tateno and Huiyu Wang and Yuchen Wang and Takuma Yagi and Mingfei Yan and Xitong Yang and Zecheng Yu and Shengxin Cindy Zha and Chen Zhao and Ziwei Zhao and Zhifan Zhu and Jeff Zhuo and Pablo Arbelaez and Gedas Bertasius and David Crandall and Dima Damen and Jakob Engel and Giovanni Maria Farinella and Antonino Furnari and Bernard Ghanem and Judy Hoffman and C. V. Jawahar and Richard Newcombe and Hyun Soo Park and James M. Rehg and Yoichi Sato and Manolis Savva and Jianbo Shi and Mike Zheng Shou and Michael Wray},
      year={2024},
      eprint={2311.18259},
      archivePrefix={arXiv},
      primaryClass={cs.CV},
      url={https://arxiv.org/abs/2311.18259}, 
}

@article{poole2022dreamfusion,
  title={Dreamfusion: Text-to-3d using 2d diffusion},
  author={Poole, Ben and Jain, Ajay and Barron, Jonathan T and Mildenhall, Ben},
  journal={arXiv preprint arXiv:2209.14988},
  year={2022}
}

@inproceedings{xie2024latte3d,
  title={Latte3d: Large-scale amortized text-to-enhanced3d synthesis},
  author={Xie, Kevin and Lorraine, Jonathan and Cao, Tianshi and Gao, Jun and Lucas, James and Torralba, Antonio and Fidler, Sanja and Zeng, Xiaohui},
  booktitle={European Conference on Computer Vision},
  pages={305--322},
  year={2024},
  organization={Springer}
}

@inproceedings{nam2024contrastive,
  title={Contrastive denoising score for text-guided latent diffusion image editing},
  author={Nam, Hyelin and Kwon, Gihyun and Park, Geon Yeong and Ye, Jong Chul},
  booktitle={Proceedings of the IEEE/CVF conference on computer vision and pattern recognition},
  pages={9192--9201},
  year={2024}
}

@inproceedings{NEURIPS2023_1a87980b,
 author = {Wang, Zhengyi and Lu, Cheng and Wang, Yikai and Bao, Fan and LI, Chongxuan and Su, Hang and Zhu, Jun},
 booktitle = {Advances in Neural Information Processing Systems},
 editor = {A. Oh and T. Naumann and A. Globerson and K. Saenko and M. Hardt and S. Levine},
 pages = {8406--8441},
 publisher = {Curran Associates, Inc.},
 title = {ProlificDreamer: High-Fidelity and Diverse Text-to-3D Generation with Variational Score Distillation},
 url = {https://proceedings.neurips.cc/paper_files/paper/2023/file/1a87980b9853e84dfb295855b425c262-Paper-Conference.pdf},
 volume = {36},
 year = {2023}
}

@inproceedings{liu2023zero,
  title={Zero-1-to-3: Zero-shot one image to 3d object},
  author={Liu, Ruoshi and Wu, Rundi and Van Hoorick, Basile and Tokmakov, Pavel and Zakharov, Sergey and Vondrick, Carl},
  booktitle={Proceedings of the IEEE/CVF international conference on computer vision},
  pages={9298--9309},
  year={2023}
}

@article{shi2023zero123++,
  title={Zero123++: a single image to consistent multi-view diffusion base model},
  author={Shi, Ruoxi and Chen, Hansheng and Zhang, Zhuoyang and Liu, Minghua and Xu, Chao and Wei, Xinyue and Chen, Linghao and Zeng, Chong and Su, Hao},
  journal={arXiv preprint arXiv:2310.15110},
  year={2023}
}

@article{liu2023one,
  title={One-2-3-45: Any single image to 3d mesh in 45 seconds without per-shape optimization},
  author={Liu, Minghua and Xu, Chao and Jin, Haian and Chen, Linghao and Varma T, Mukund and Xu, Zexiang and Su, Hao},
  journal={Advances in Neural Information Processing Systems},
  volume={36},
  pages={22226--22246},
  year={2023}
}

@inproceedings{liu2024one,
  title={One-2-3-45++: Fast single image to 3d objects with consistent multi-view generation and 3d diffusion},
  author={Liu, Minghua and Shi, Ruoxi and Chen, Linghao and Zhang, Zhuoyang and Xu, Chao and Wei, Xinyue and Chen, Hansheng and Zeng, Chong and Gu, Jiayuan and Su, Hao},
  booktitle={Proceedings of the IEEE/CVF conference on computer vision and pattern recognition},
  pages={10072--10083},
  year={2024}
}

@article{gao2024cat3d,
  title={Cat3d: Create anything in 3d with multi-view diffusion models},
  author={Gao, Ruiqi and Holynski, Aleksander and Henzler, Philipp and Brussee, Arthur and Martin-Brualla, Ricardo and Srinivasan, Pratul and Barron, Jonathan T and Poole, Ben},
  journal={arXiv preprint arXiv:2405.10314},
  year={2024}
}

@misc{sobol2024zerotoheroenhancingzeroshotnovel,
      title={Zero-to-Hero: Enhancing Zero-Shot Novel View Synthesis via Attention Map Filtering}, 
      author={Ido Sobol and Chenfeng Xu and Or Litany},
      year={2024},
      eprint={2405.18677},
      archivePrefix={arXiv},
      primaryClass={cs.CV},
      url={https://arxiv.org/abs/2405.18677}, 
}

@misc{alliegro2023polydiffgenerating3dpolygonal,
      title={PolyDiff: Generating 3D Polygonal Meshes with Diffusion Models}, 
      author={Antonio Alliegro and Yawar Siddiqui and Tatiana Tommasi and Matthias Nießner},
      year={2023},
      eprint={2312.11417},
      archivePrefix={arXiv},
      primaryClass={cs.CV},
      url={https://arxiv.org/abs/2312.11417}, 
}

@misc{liu2023meshdiffusionscorebasedgenerative3d,
      title={MeshDiffusion: Score-based Generative 3D Mesh Modeling}, 
      author={Zhen Liu and Yao Feng and Michael J. Black and Derek Nowrouzezahrai and Liam Paull and Weiyang Liu},
      year={2023},
      eprint={2303.08133},
      archivePrefix={arXiv},
      primaryClass={cs.GR},
      url={https://arxiv.org/abs/2303.08133}, 
}

@inproceedings{10.1145/3680528.3687699,
author = {Roessle, Barbara and M\"{u}ller, Norman and Porzi, Lorenzo and Rota Bul\`{o}, Samuel and Kontschieder, Peter and Dai, Angela and Nie\ss{}ner, Matthias},
title = {L3DG: Latent 3D Gaussian Diffusion},
year = {2024},
isbn = {9798400711312},
publisher = {Association for Computing Machinery},
address = {New York, NY, USA},
url = {https://doi.org/10.1145/3680528.3687699},
doi = {10.1145/3680528.3687699},
booktitle = {SIGGRAPH Asia 2024 Conference Papers},
articleno = {37},
numpages = {11},
location = {Tokyo, Japan},
series = {SA '24}
}

@misc{mu2024gsdviewguidedgaussiansplatting,
      title={GSD: View-Guided Gaussian Splatting Diffusion for 3D Reconstruction}, 
      author={Yuxuan Mu and Xinxin Zuo and Chuan Guo and Yilin Wang and Juwei Lu and Xiaofeng Wu and Songcen Xu and Peng Dai and Youliang Yan and Li Cheng},
      year={2024},
      eprint={2407.04237},
      archivePrefix={arXiv},
      primaryClass={cs.CV},
      url={https://arxiv.org/abs/2407.04237}, 
}

@inproceedings{zeng2022lion,
    title={LION: Latent Point Diffusion Models for 3D Shape Generation},
    author={Xiaohui Zeng and Arash Vahdat and Francis Williams and Zan Gojcic and Or Litany and Sanja Fidler and Karsten Kreis},
    booktitle={Advances in Neural Information Processing Systems (NeurIPS)},
    year={2022}
}

@misc{jiang2023rtmposerealtimemultipersonpose,
      title={RTMPose: Real-Time Multi-Person Pose Estimation based on MMPose}, 
      author={Tao Jiang and Peng Lu and Li Zhang and Ningsheng Ma and Rui Han and Chengqi Lyu and Yining Li and Kai Chen},
      year={2023},
      eprint={2303.07399},
      archivePrefix={arXiv},
      primaryClass={cs.CV},
      url={https://arxiv.org/abs/2303.07399}, 
}

@misc{cao2019openposerealtimemultiperson2d,
      title={OpenPose: Realtime Multi-Person 2D Pose Estimation using Part Affinity Fields}, 
      author={Zhe Cao and Gines Hidalgo and Tomas Simon and Shih-En Wei and Yaser Sheikh},
      year={2019},
      eprint={1812.08008},
      archivePrefix={arXiv},
      primaryClass={cs.CV},
      url={https://arxiv.org/abs/1812.08008}, 
}

@article{10.1109/TPAMI.2022.3222784,
author = {Fang, Hao-Shu and Li, Jiefeng and Tang, Hongyang and Xu, Chao and Zhu, Haoyi and Xiu, Yuliang and Li, Yong-Lu and Lu, Cewu},
title = {AlphaPose: Whole-Body Regional Multi-Person Pose Estimation and Tracking in Real-Time},
year = {2023},
issue_date = {June 2023},
publisher = {IEEE Computer Society},
address = {USA},
volume = {45},
number = {6},
issn = {0162-8828},
url = {https://doi.org/10.1109/TPAMI.2022.3222784},
doi = {10.1109/TPAMI.2022.3222784},
journal = {IEEE Trans. Pattern Anal. Mach. Intell.},
month = jun,
pages = {7157–7173},
numpages = {17}
}

@misc{wang2020deephighresolutionrepresentationlearning,
      title={Deep High-Resolution Representation Learning for Visual Recognition}, 
      author={Jingdong Wang and Ke Sun and Tianheng Cheng and Borui Jiang and Chaorui Deng and Yang Zhao and Dong Liu and Yadong Mu and Mingkui Tan and Xinggang Wang and Wenyu Liu and Bin Xiao},
      year={2020},
      eprint={1908.07919},
      archivePrefix={arXiv},
      primaryClass={cs.CV},
      url={https://arxiv.org/abs/1908.07919}, 
}

@article{doi:10.1177/027836498600500404,
author = {Randall C. Smith and Peter Cheeseman},
title ={On the Representation and Estimation of Spatial Uncertainty},

journal = {The International Journal of Robotics Research},
volume = {5},
number = {4},
pages = {56-68},
year = {1986},
doi = {10.1177/027836498600500404},

URL = {https://doi.org/10.1177/027836498600500404},
eprint = {https://doi.org/10.1177/027836498600500404},
}

@Article{robotics11010024,
AUTHOR = {Macario Barros, Andréa and Michel, Maugan and Moline, Yoann and Corre, Gwenolé and Carrel, Frédérick},
TITLE = {A Comprehensive Survey of Visual SLAM Algorithms},
JOURNAL = {Robotics},
VOLUME = {11},
YEAR = {2022},
NUMBER = {1},
ARTICLE-NUMBER = {24},
URL = {https://www.mdpi.com/2218-6581/11/1/24},
ISSN = {2218-6581},
DOI = {10.3390/robotics11010024}
}

@article{Shi_2020,
   title={MotioNet: 3D Human Motion Reconstruction from Monocular Video with Skeleton Consistency},
   volume={40},
   ISSN={1557-7368},
   url={http://dx.doi.org/10.1145/3407659},
   DOI={10.1145/3407659},
   number={1},
   journal={ACM Transactions on Graphics},
   publisher={Association for Computing Machinery (ACM)},
   author={Shi, Mingyi and Aberman, Kfir and Aristidou, Andreas and Komura, Taku and Lischinski, Dani and Cohen-Or, Daniel and Chen, Baoquan},
   year={2020},
   month=sep, pages={1–15} }

@inproceedings{oord2017vqvae,
author = {van den Oord, Aaron and Vinyals, Oriol and Kavukcuoglu, Koray},
title = {Neural discrete representation learning},
year = {2017},
isbn = {9781510860964},
publisher = {Curran Associates Inc.},
address = {Red Hook, NY, USA},
booktitle = {Proceedings of the 31st International Conference on Neural Information Processing Systems},
pages = {6309–6318},
numpages = {10},
location = {Long Beach, California, USA},
series = {NIPS'17}
}

@article{Kerbl2023_3DGS,
author = {Kerbl, Bernhard and Kopanas, Georgios and Leimkuehler, Thomas and Drettakis, George},
title = {3D Gaussian Splatting for Real-Time Radiance Field Rendering},
year = {2023},
issue_date = {August 2023},
publisher = {Association for Computing Machinery},
address = {New York, NY, USA},
volume = {42},
number = {4},
issn = {0730-0301},
url = {https://doi.org/10.1145/3592433},
doi = {10.1145/3592433},
journal = {ACM Trans. Graph.},
month = jul,
articleno = {139},
numpages = {14}
}

@inproceedings{2023ObjaverseXL,
author = {Deitke, Matt and Liu, Ruoshi and Wallingford, Matthew and Ngo, Huong and Michel, Oscar and Kusupati, Aditya and Fan, Alan and Laforte, Christian and Voleti, Vikram and Gadre, Samir Yitzhak and VanderBilt, Eli and Kembhavi, Aniruddha and Vondrick, Carl and Gkioxari, Georgia and Ehsani, Kiana and Schmidt, Ludwig and Farhadi, Ali},
title = {Objaverse-XL: a universe of 10M+ 3D objects},
year = {2023},
publisher = {Curran Associates Inc.},
address = {Red Hook, NY, USA},
booktitle = {Proceedings of the 37th International Conference on Neural Information Processing Systems},
articleno = {1554},
numpages = {15},
location = {New Orleans, LA, USA},
series = {NIPS '23}
}

@article{genmo2025,
  title={GENMO: Generative Models for Human Motion Synthesis},
  author={Li, Jiefeng and Cao, Jinkun and Zhang, Haotian and Rempe, Davis and Kautz, Jan and Iqbal, Umar and Yuan, Ye},
  journal={arXiv preprint arXiv:2505.01425},
  year={2025}
}

@article{zhang2024large,
      title   =   {Large Motion Model for Unified Multi-Modal Motion Generation}, 
      author  =   {Zhang, Mingyuan and Jin, Daisheng and Gu, Chenyang and Hong, Fangzhou and Cai, Zhongang and Huang, Jingfang and Zhang, Chongzhi and Guo, Xinying and Yang, Lei and He, Ying and Liu, Ziwei},
      year    =   {2024},
      journal =   {arXiv preprint arXiv:2404.01284},
}

@misc{song2021scorebasedgenerativemodelingstochastic,
      title={Score-Based Generative Modeling through Stochastic Differential Equations}, 
      author={Yang Song and Jascha Sohl-Dickstein and Diederik P. Kingma and Abhishek Kumar and Stefano Ermon and Ben Poole},
      year={2021},
      eprint={2011.13456},
      archivePrefix={arXiv},
      primaryClass={cs.LG},
      url={https://arxiv.org/abs/2011.13456}, 
}

@article{song2020denoising,
  title={Denoising diffusion implicit models},
  author={Song, Jiaming and Meng, Chenlin and Ermon, Stefano},
  journal={arXiv preprint arXiv:2010.02502},
  year={2020}
}

@article{Kynkaanniemi2019,
  author    = {Tuomas Kynkäänniemi and Tero Karras
               and Samuli Laine and Jaakko Lehtinen
               and Timo Aila},
  title     = {Improved Precision and Recall Metric for Assessing Generative Models},
  journal   = {CoRR},
  volume    = {abs/1904.06991},
  year      = {2019},
}

@inproceedings{binkowski2018KID,
title={Demystifying {MMD} {GAN}s},
author={Mikołaj Bińkowski and Dougal J. Sutherland and Michael Arbel and Arthur Gretton},
booktitle={International Conference on Learning Representations},
year={2018},
url={https://openreview.net/forum?id=r1lUOzWCW},
}

@article{opencv_library,
    author = {Bradski, G.},
    journal = {Dr. Dobb's Journal of Software Tools},
    title = {{The OpenCV Library}},
    year = {2000}
}
\bibliographystyle{abbrvnat}

\newpage
\appendix
\setcounter{page}{1}

\clearpage

\appendix

\section{Weights for 3D to 2D Loss Equivalence}
\label{app:proof}

In standard DDPM and DDIM training, given a sample $\mathbf{x} \sim p$ and denoiser output $\hat{\mathbf{x}}$, the reconstruction loss is the mean squared error:
\[
\mathbb{L}_{3d} = \mathbb{E}_{\mathbf{x}\sim p}\bigl[\lVert \hat{\mathbf{x}} - \mathbf{x}\rVert_2^2 \bigr].
\]

Since the loss decomposes over coordinates, we ignore any additional structure (e.g., joint  $J$ or frame $F$) and focus on a single 3D point:
\[
\mathbf{x} =
\begin{bmatrix}
x \\ y \\ z
\end{bmatrix},\;\;
\hat{\mathbf{x}} =
\begin{bmatrix}
\hat{x} \\ \hat{y} \\ \hat{z}
\end{bmatrix}.
\]

For 2D projection, let $\psi$ denote the camera elevation angle and $\theta$ the azimuth angle. We denote by $\mathcal{P}(\mathbf{x},\psi,\theta)$ the perspective projection of $\mathbf{x}$ onto the camera image plane:
\[
\mathcal{P}(\mathbf{x},\psi,\theta)=
\begin{bmatrix}
u \\ v
\end{bmatrix}
=
\begin{bmatrix}
\frac{\cos\theta\cos\psi\,x + \sin\theta\cos\psi\,z}{d(\mathbf{x},\theta,\psi)} \\
\frac{\cos\psi\,y + \cos\theta\sin\psi\,x + \sin\theta\sin\psi\,z}{d(\mathbf{x},\theta,\psi)}
\end{bmatrix},
\]
with
\[
d(\mathbf{x},\theta,\psi)
=
\sin\psi\,y + \cos\theta\cos\psi\,x + \sin\theta\cos\psi\,z.
\]

We define the 2D MSE loss as
\[
\mathbb{L}_{2d}
=
\mathbb{E}_{\mathbf{x}\sim p,\;\theta\sim\mathcal{U}[0,2\pi]}
\!\left[
\left\| W \odot
\bigl(\mathcal{P}(\hat{\mathbf{x}},\psi,\theta)
-
\mathcal{P}(\mathbf{x},\psi,\theta)\bigr)
\right\|_2^2
\right],
\]
where $W = [W_u, W_v]^\top$ contains per-axis weights and $\odot$ denotes element-wise multiplication.

\textbf{Depth assumption.}  
We approximate the projection denominator as constant between $\mathbf{x}$ and $\hat{\mathbf{x}}$:
\[
d \triangleq d(\mathbf{x},\theta,\psi)
=
d(\hat{\mathbf{x}},\theta,\psi)
\]

We now show that there exist weights $W_u,W_v$ such that $\mathbb{L}_{2d} = \mathbb{L}_{3d}$, and that they are both proportional to $d$, the depth of the point in camera coordinates.

\textbf{Proof.}

Consider the normalization weights
\[
W_u = \frac{d}{\Phi},
\qquad
W_v = \frac{d}{\cos\psi},
\qquad
\Phi = \frac{\cos\psi}{\sqrt{\,2 - \tan^2\psi\,}}.
\]

\textbf{Image $v$-axis contribution.}
{
\begin{align*}
&\frac{1}{2\pi}\int_{0}^{2\pi}
\left(
\frac{d}{\cos\psi}
\left(
\frac{\cos\psi\,y + \cos\theta\sin\psi\,x + \sin\theta\sin\psi\,z}{d}
-
\frac{\cos\psi\,\hat{y} + \cos\theta\sin\psi\,\hat{x} + \sin\theta\sin\psi\,\hat{z}}{d}
\right)
\right)^{2}\, d\theta
\\
&=
\frac{1}{2\pi}\int_{0}^{2\pi}
\left(
y + \cos\theta\tan\psi\,x + \sin\theta\tan\psi\,z
- \hat{y} - \cos\theta\tan\psi\,\hat{x} - \sin\theta\tan\psi\,\hat{z}
\right)^{2}\, d\theta
\\
&=
\frac{1}{2\pi}\int_{0}^{2\pi}
\Bigl(
2\cos\theta\tan\psi\,(yx - \hat{y}x - y\hat{x} + \hat{y}\hat{x})
+ 2\sin\theta\tan\psi\,(yz - \hat{y}z - y\hat{z} + \hat{y}\hat{z})
\\[-2pt]
&\quad\quad
+ 2\sin\theta\cos\theta\tan^{2}\psi\,(xz - \hat{x}z - x\hat{z} + \hat{x}\hat{z})
+ (y^{2} - 2y\hat{y} + \hat{y}^{2})
\\[-2pt]
&\quad\quad
+ \cos^{2}\theta\tan^{2}\psi\,(x^{2} - 2x\hat{x} + \hat{x}^{2})
+ \sin^{2}\theta\tan^{2}\psi\,(z^{2} - 2z\hat{z} + \hat{z}^{2})
\Bigr)\, d\theta
\\
&\overset{(1)}{=}
(y - \hat{y})^{2}
+ \tfrac{1}{2}\tan^{2}\psi\bigl((x - \hat{x})^{2} + (z - \hat{z})^{2}\bigr),
\end{align*}
where step $(1)$ uses linearity of integration and the identities
\begin{align*}
\int_{0}^{2\pi}\cos\theta\,d\theta
= \int_{0}^{2\pi}\sin\theta\,d\theta
= \int_{0}^{2\pi}\cos\theta\sin\theta\,d\theta
&= 0, \\
\int_{0}^{2\pi}\cos^{2}\theta\,d\theta
= \int_{0}^{2\pi}\sin^{2}\theta\,d\theta
&= \pi.
\end{align*}

\textbf{Image $u$-axis contribution.}
\begin{align*}
&\frac{1}{2\pi}\int_{0}^{2\pi}
\left(
\frac{d}{\Phi}
\left(
\frac{\cos\theta\cos\psi\,x + \sin\theta\cos\psi\,z}{d}
-
\frac{\cos\theta\cos\psi\,\hat{x} + \sin\theta\cos\psi\,\hat{z}}{d}
\right)
\right)^{2}\, d\theta
\\
&=
\frac{1}{2\pi}\int_{0}^{2\pi}
\left(
\cos\theta\frac{\cos\psi}{\Phi}(x-\hat{x})
+ \sin\theta\frac{\cos\psi}{\Phi}(z-\hat{z})
\right)^{2}\, d\theta
\\
&=
\frac{\cos^{2}\psi}{2\pi\Phi^{2}}
\int_{0}^{2\pi}
\Bigl(
\cos^{2}\theta\,(x-\hat{x})^{2}
+ 2\sin\theta\cos\theta\,(x-\hat{x})(z-\hat{z})
+ \sin^{2}\theta\,(z-\hat{z})^{2}
\Bigr)\, d\theta
\\
&\overset{(1)}{=}
\frac{\cos^{2}\psi}{2\Phi^{2}}\bigl((x-\hat{x})^{2} + (z-\hat{z})^{2}\bigr).
\end{align*}

Summing the coefficients on $(x-\hat{x})^{2} + (z-\hat{z})^{2}$ from the $u$ and $v$ contributions:
\begin{align*}
\tfrac{1}{2}\tan^{2}\psi
+ \frac{\cos^{2}\psi}{2\Phi^{2}}
&=
\tfrac{1}{2}\tan^{2}\psi
+ \frac{\cos^{2}\psi}{2\left(\frac{\cos\psi}{\sqrt{2-\tan^{2}\psi}}\right)^{2}}
=
\tfrac{1}{2}\tan^{2}\psi + \frac{2 - \tan^{2}\psi}{2} = 1.
\end{align*}
}

Thus, for every sample,
\[
\mathbb{E}_{\theta}
\!\left[
\left\|W \odot (\mathcal{P}(\hat{\mathbf{x}},\psi,\theta) - \mathcal{P}(\mathbf{x},\psi,\theta))\right\|_2^2
\right]
=
\|\hat{\mathbf{x}} - \mathbf{x}\|_2^2,
\]
and therefore $\mathbb{L}_{2d} = \mathbb{L}_{3d}$.

\section{Projection of a 3D Point onto a 2D Camera Ray}
\label{app:ray_projecting}
We consider a calibrated pinhole camera $\Pi$ with extrinsics $[\,\mathbf{R}\,|\,\mathbf{t}\,] \in \mathbb{R}^{3\times 4}$ that map world points to the camera frame as
\[
\mathbf{x}_{\mathrm{cam}} = \mathbf{R}\,\mathbf{x}_{\mathrm{world}} + \mathbf{t}, 
\qquad
\mathbf{R}\in\mathrm{SO}(3),\ \mathbf{t}\in\mathbb{R}^3.
\]
Let $y=(u,v)^T\in \mathbb{R}^2$ denote image coordinates in normalized units (after applying the inverse of the intrinsics, $K^{-1}$), and define the camera-frame ray direction
\[
\mathbf{y_H} \;=\; \begin{bmatrix} u\\[2pt] v\\[2pt] 1 \end{bmatrix} \in \mathbb{R}^3.
\]
\noindent\textbf{World ray.}
The camera center in world coordinates and the corresponding world-frame ray direction are
\begin{equation}
    \mathbf{C} \;=\; -\,\mathbf{R}^{\!\top}\mathbf{t}, 
    \qquad
    \mathbf{r} \;=\; \mathbf{R}^{\!\top}\mathbf{y_H}.
\end{equation}
Thus, the camera ray in world coordinates is the line
\[
\ell(\lambda) \;=\; \mathbf{C} + \lambda\,\mathbf{r}, \qquad \lambda \in \mathbb{R}.
\]

\vspace{4pt}
\noindent\textbf{Orthogonal projection onto the ray.}
Given a world point $\mathbf{x}\in\mathbb{R}^3$, its closest point $\mathbf{P}_\Pi$ on the ray $\ell$ is obtained by minimizing
\(
\|\mathbf{C} + \lambda \mathbf{r} - \mathbf{x}\|_2^2
\)
with respect to $\lambda$. The optimal scalar is
\begin{equation}
    \lambda^\star \;=\; \frac{\mathbf{r}^{\!\top}(\mathbf{x}-\mathbf{C})}{\mathbf{r}^{\!\top}\mathbf{r}}
    \;=\; \frac{\mathbf{r}^{\!\top}\mathbf{x} + \mathbf{y_H}^{\!\top}\mathbf{t}}{\mathbf{y_H}^{\!\top}\mathbf{y_H}},
    \label{eq:lambda}
\end{equation}
where the second equality uses $\mathbf{C}=-\mathbf{R}^{\!\top}\mathbf{t}$ and the orthogonality of $\mathbf{R}$:
\(
\mathbf{r}^{\!\top}\mathbf{r}
= (\mathbf{R}^{\!\top}\mathbf{d})^{\!\top}(\mathbf{R}^{\!\top}\mathbf{d})
= \mathbf{d}^{\!\top}\mathbf{d}.
\)
The projected point is then
\begin{equation}
    \mathbf{P_\Pi(x, y)} \;=\; \mathbf{C} + \lambda^\star \mathbf{r}.
\end{equation}

In the paper, we use $P_\Pi(\mathbf{\hat x_0}, \mathbf{y})$ to denote the above operation, applied separately per frame and per joint.

\section{Experiments Technical Details}\label{app:exp_tech_details}
\subsection{HumanML3D Data Processing}
\label{app:dataprocessing_humanML}
We construct a 2D-only version of HumanML3D by sampling, for each motion, a camera with azimuth $\sim \mathcal{U}[-\pi, \pi]$, elevation $\sim \mathcal{U}[0, \pi/8]$, and position constrained so that the closest joint is at least 3 units from the camera. We project the 3D motion to 2D and discard the 3D ground truth from the training set. To obtain the noisy 3D teacher required by our method, we lift the 2D motions back to 3D using either MotionBERT~\cite{motionbert2022} or MVLift~\cite{li2025mvlift}. MotionBERT produces centered poses, so we recover an uncentered trajectory by estimating root depth from observed limb lengths (\cref{app:naive_depth}); MVLift predicts global root trajectories directly and requires no such recovery.

\paragraph{Skeleton Conversion}
Both motionBERT and MVLift require a different skeleton than HumanML3D, in order to be as fair to the baselines as possible, we used HumanML3D data to render SMPL meshes, for which vertices conversion to the skeletons format is known, and convert the joints in 3D, and then rendered them to 2D for the lifter to lift using the same camera.

\subsection{HumanML3D Evaluation Metrics}
\label{app:evalmetrics_humanML}
We report FID (statistical similarity between real and generated motion features), Diversity (average distance between random pairs), R-Precision Top-3 (text--motion alignment in a shared embedding), Multimodal Distance (distance to ground-truth motions), and Multimodality (variance across samples generated from the same prompt). All results are averaged over 20 generations and reported with 95\% confidence intervals.

\subsection{Fit3D Data Processing}
\label{app:dataprocessing_fit3D}
For each 2D sequence extracted by RTMPose~\cite{jiang2023rtmposerealtimemultipersonpose} BodyWithFeet performance model, we further smooth it \amir{using weighted temporal mean (with weights (0.25, 0.5, 0.25)}, and convert to the SMPL joint format using manually picked weight. Sequences are split into clips matching the HumanML3D length range using the dynamic-programming procedure \amir{detailed in} \cref{app:clip_splitting}. Text descriptions are generated from the exercise names in HumanML3D's natural-language style.

For WHAM we use the official implementation and checkpoint, cut motions to clip length, and align them to the dataset frame on the first frame of each clip to avoid penalizing WHAM for not using the provided cameras at lifting evaluation time.

\paragraph{Skeleton Conversion}
WHAM requires no skeleton conversion as it uses the raw RGB image and outputs in SMPL format. MVLift, however, requires 17-keypoint skeletons. Since all work on fit3D that relies on 2D pose estimation uses RTMPose as pseudo-GT 2D poses, we needed to convert the joints directly in 2D to the MVLift skeleton format. To do so, we trained a 2D joint regressor using the accurate paired data created on HumanML3D, converting from the 22-joint format (HumanML) to the 17-joint format (MVLift), and applied it to all 2D poses.

\subsection{Clip Extraction for Fit3D}
\label{app:clip_splitting}

The raw sequences in the dataset \cite{Fieraru_2021_CVPR} vary significantly in duration, so we split each motion into shorter, natural clips using a lightweight dynamic-programming procedure. The goal is to produce segments that (i) fall within a desired length range, (ii) avoid splitting at high-motion frames, and (iii) prefer frames whose pose is close to the rest pose.

\paragraph{Split Cost.}
For each candidate split frame $i$, we compute a frame cost
\[
\ell_{\text{frame}}(i)
= \lambda_{\text{pose}}\,\ell_{\text{pose}}(i)
+ \lambda_{\text{vel}}\,\ell_{\text{vel}}(i),
\]
where $\ell_{\text{pose}}(i)$ measures the MSE between the local pose at frame $i$ and the rest local pose, and $\ell_{\text{vel}}(i)$ is the average local velocity magnitude around frame $i$. We use $\lambda_{\text{pose}}=30000$ and $\lambda_{\text{vel}}=15000$.

We also apply a clip-length penalty
\[
\ell_{\text{len}}(\Delta) = (\Delta - L_{\text{target}})^2,
\]
with $L_{\min}=60$, $L_{\text{target}}=150$, and $L_{\max}=300$.

\paragraph{Dynamic Programming.}
Let $\mathrm{DP}[i]$ store the best cumulative cost for segmenting frames $[0,i]$ and the index of the previous split. For each frame $i$, we consider all valid previous splits $j$ such that $L_{\min} \le i-j \le L_{\max}$, and select the $j$ minimizing
\[
\mathrm{DP}[j] 
+ \ell_{\text{len}}(i-j) 
+ \ell_{\text{frame}}(i).
\]
After the forward pass, we choose the best terminal split and backtrack through the stored indices to recover all segment boundaries.

\begin{figure}[t]
    \centering
    \includegraphics[width=\linewidth]{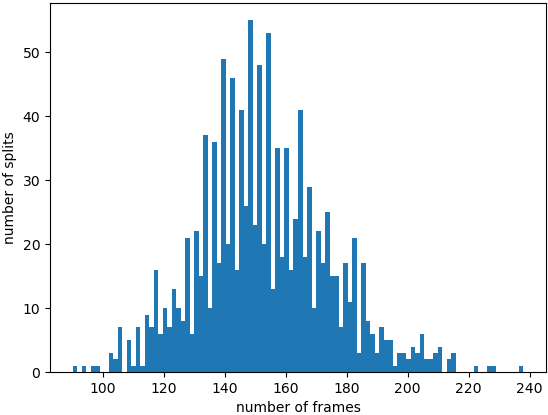}
    \caption{Histogram of resulting clip lengths after dynamic-programming segmentation.}
    \label{fig:clip_length_hist}
\end{figure}

\paragraph{Outcome.}
This produces stable, naturally aligned clips that avoid erratic split points while keeping durations consistent across the dataset. Running the procedure over all sequences yields 1{,}161 clips in total (including subject $s09$). See \cref{fig:clip_length_hist}.

\subsection{NBA Data Processing}
Because the motions are scaled and centered, recovering a real pinhole camera from the data is infeasible. We therefore adopt MAS's convention: a camera 7 units along the $Z$-axis at $\pi/16$ elevation. ElePose serves as the noisy teacher.

\subsection{Compute Cost Estimates}
\label{app:compute}

All experiments were run on a single NVIDIA RTX 4090 GPU (24\,GB VRAM).
Table~\ref{tab:compute} summarises the compute required for each method.
Lifter preprocessing (running MotionBERT or MVLift on the full dataset) is a one-time cost shared by any lifter-based approach; the resulting pseudo-3D labels are reused across all training runs.

\begin{table}[h]
\centering
\caption{Compute costs on a single RTX 4090. Inference is reported in seconds per 100 generated samples.}
\label{tab:compute}
\begin{tabular}{lccc}
\toprule
\textbf{Stage} & \textbf{Ours} & \textbf{MDM (on lifter)} & \textbf{MAS} \\
\midrule
Lifter preprocessing (hr) & $\sim$200 & $\sim$200 & --- \\
Training (hr)              & $\sim$46  & $\sim$36  & $\sim$6 \\
Inference (sec / 100)      & $\sim$5   & $\sim$5   & $\sim$12 \\
\bottomrule
\end{tabular}
\end{table}

\paragraph{Hyperparameters Search}
The sweep described in Appendix~\ref{app:hyperparameters} comprised approximately 56 runs of $\sim$20 hours each ($\sim$1{,}120 GPU-hours total), which is not reflected in the per-method figures above.

\paragraph{Earlier Exploration}
Besides the compute presented here, earlier versions of the paper with contributions which were removed because of their lack of support from the ablations included other hyper parameter searches, one small and one larger, as well as early exploration training and the blender rendering of all motions for the survey / supplementary material. We do not have an exact number for total compute used but it could be estimated as around $2.5\times$ the values reported for the hyper parameter search and lifting / training models for experiments.
\section{Naive Root Depth Estimation}
\label{app:naive_depth}

We denote by $\{ \mathbf{x}_i^{3D} \}_{i=1}^{22}$ the 3D joint locations of the articulated skeleton and by 
$\{ \mathbf{x}_i^{2D} \}_{i=1}^{22}$ their image projections under a calibrated pinhole camera with focal length~$f$.
The skeleton connectivity forms a tree whose edges (bones) have known rest lengths
$L^{3D}_i = \| \mathbf{x}^{3D}_{p(i)} - \mathbf{x}^{3D}_i \|$,
where $p(i)$ is the parent joint of node~$i$.
For each frame, we observe only the 2D joint coordinates and the corresponding
2D bone lengths
$L^{2D}_i = \| \mathbf{x}^{2D}_{p(i)} - \mathbf{x}^{2D}_i \|$.

\vspace{4pt}
\noindent\textbf{Assumptions.}
We (falsely) assume that the global 3D orientation of the skeleton is uniformly random,
and that the three bones incident to the root joint (pelvis)
are independently and uniformly oriented in~$\mathrm{SO}(3)$.
Under this assumption, the direction of each bone is uniformly distributed
over the unit sphere, so that $\cos\theta \sim \mathrm{Unif}[-1,1]$,
where $\theta$ is the angle between the bone and the camera optical axis.

For a bone of true length $L^{3D}$ viewed at mean depth~$z$
and at an angle~$\theta$ to the optical axis,
its projected 2D length satisfies approximately
\begin{equation}
    L^{2D} \approx \frac{f}{z} L^{3D} \sin\theta,
\end{equation}
which rearranges to
\begin{equation}
    \frac{L^{3D}}{L^{2D}} \approx \frac{z}{f\,\sin\theta}.
\end{equation}

\vspace{4pt}
\noindent\textbf{Estimator.}
Because the orientations are random, with high probability at least one of the
three root-adjacent bones will have $\sin\theta$ close to~1.
We therefore define a simple, ``naive'' depth estimator as
\begin{equation}
    \hat{z}_{\text{root}}
    = f\,C_{3}\,\min_{i \in \mathcal{N}_{\text{root}}}
      \frac{L^{3D}_i}{L^{2D}_i},
\end{equation}
where $\mathcal{N}_{\text{root}}$ denotes the three edges connected to the root.

\vspace{4pt}
\noindent\textbf{Constant.}
The constant $C_{3}$ corrects the expectation bias due to the
non-uniform distribution of $\sin\theta$ under uniform random orientations.
By integrating over this distribution for three independent samples,
we obtain
\begin{equation}
    C_{3} = \frac{1}{\mathbb{E}[1/S_{\max}]}
          \approx 0.9358,
\end{equation}
where $S_{\max} = \max(\sin\theta_1,\sin\theta_2,\sin\theta_3)$.
Thus, in normalized camera coordinates ($f=1$),
\[
    \hat{z}_{\text{root}} \approx 0.9358
    \min_{i \in \mathcal{N}_{\text{root}}} \frac{L^{3D}_i}{L^{2D}_i}.
\]

\vspace{4pt}
\noindent
This estimator provides a simple and numerically stable baseline
for root-depth inference from 2D skeletons,
assuming uniformly random joint orientations and known bone lengths.

\section{Formal Definition of Recall$^\dagger$}
\label{app:nba_extra}

\paragraph{Recall Alternative.}
\cite{Kynkaanniemi2019} define the notion of "close enough" as follows: for each sample $\phi’$, find its $K-$th (usually taken as 3) nearest neighbor within its own manifold $\Phi$, and say that sample $\phi$ is "close enough" if $|\phi - \phi’|_2 \leq |\phi’ - NN_K(\phi’, \Phi)|_2$.
Thus, for real and generated features $\Phi_r, \Phi_g$, we can define:
\[
P = \frac{1}{|\Phi_g|}\sum_{\phi \in \Phi_g} \mathbf{1}\{\exists \phi’ \in \Phi_r : |\phi - \phi’|_2 \leq |\phi’ - NN_K(\phi’, \Phi_r)|_2\}
\]
\[
R = \frac{1}{|\Phi_r|}\sum_{\phi \in \Phi_r} \mathbf{1}\{\exists \phi’ \in \Phi_g : |\phi - \phi’|_2 \leq |\phi’ - NN_K(\phi’, \Phi_g)|_2\}
\]
Note that to estimate recall, we open hyperspheres around generated samples based on \textit{generated} nearest neighbors. This biases the metric towards favoring exaggerated diversity in the generated distribution. We therefore suggest a variant of the recall metric that respects the spread of the real distribution by computing the NN in the real distribution.
\[
R^\dagger = \frac{1}{|\Phi_r|}\sum_{\phi \in \Phi_r} \mathbf{1}\{\exists \phi’ \in \Phi_g : |\phi - \phi’|_2 \leq |\phi - NN_K(\phi, \Phi_r)|_2\}
\]

\section{Hyper Parameter Choices}
\label{app:hyperparameters}

\paragraph{Overview.}
The only hyperparameters shown to affect method quality were $\lambda_{vel}$ $t^*$ and number of steps for multistep under $t^*$. The hyper parameter search resulted in $\lambda_{vel}=287$ on HumanML3D and $\lambda_{vel}=187$ on NBA. $t^*=30$ on HumanML3D and $t^*=23$ on NBA and $\#steps=3$ on HumanML3D and $\#steps=4$ on NBA.
\paragraph{MDM Hyperparameters.}
For MDM, we follow the original configuration from \cite{tevet2023human}, with two exceptions:
(i) we use $T=50$ diffusion steps, which has become standard practice over the past year, and  
(ii) for the NBA experiment, we modify the noise schedule by setting $\tau = 2$ to place greater emphasis on high-noise denoising stages.

\section{Mathematical Formulations of Diffusion Sampling and Guidance}
\label{app:formal_guidance}

We assume the DDIM notation of \cite{song2020denoising}, where for a noise schedule $t \in \{1,\ldots,T\}$ with noising weights $\beta_1,\ldots,\beta_T$, the cumulative product $\alpha_t=\prod_{j=1}^{t} (1-\beta_j)$ defines the forward noising process of data $\mathbf{x}\sim p$:
\[
\begin{aligned}
\mathbf{x} &\sim p \\
q(\mathbf{x}_{t} \mid \mathbf{x}) &\sim \mathcal{N}(\sqrt{1-\alpha_t}\,\mathbf{x},\, \alpha_t \mathbf{I}) \\
\mathbf{x}_T &\sim \mathcal{N}(0,\mathbf{I}) \\
\end{aligned}
\]
DDPM~\cite{ho2020denoisingdiffusionprobabilisticmodels} provides a method to sample from $p$ using a denoiser that estimates $p(\mathbf{x}\mid \mathbf{x}_t)$, as summarized in \cref{par:DDPM_sampling}. DDIM generalizes this formulation and enables more efficient sampling, described (in part) in \cref{par:multistep_training}.

\subsection{DDIM Used in Training}

\paragraph{Formal Multistep Training.}
\label{par:multistep_training}
During training, to optimize the model over intermediate steps $t < t^*$, we use DDIM sampling to skip from some $t_b \ge t^*$ down to $t_{b-1},\ldots,t_1$, where $b$ is the number of buckets defined by LIS~\cite{peng2025lesson}. Each $t_i$ is sampled uniformly as $t_i \sim \mathcal{U}((i-1)B,\,(i-1)B + B - 1)$, with bucket size $B=\lceil t^*/(b-1) \rceil$.  
For our hyperparameters $t^*=12$ and $b=3$, this yields  
$t_3\sim\mathcal{U}(12,17)$, $t_2\sim\mathcal{U}(6,11)$, and $t_1\sim\mathcal{U}(0,5)$.

We randomly mix batches where $t \ge t^*$ and the standard diffusion loss is used, with batches where $t < t^*$. For the latter, we first sample  
$\mathbf{x}_{t_3}=\sqrt{\alpha_{t_3}}\,\tilde{\mathbf{x}} + \sqrt{1-\alpha_{t_3}}\epsilon$  
with $\epsilon \sim \mathcal{N}(0,\mathbf{I})$.  
We then iteratively compute, for $i = 3,2,1$:
\begin{align}
\hat{\mathbf{x}}_{i\rightarrow0} &= D_\theta(\text{stop\_gradient}(\mathbf{x}_{t_i}),\, t_i) \\
\mathcal{L}_i &= \mathcal{L}_{\text{total}}(\hat{\mathbf{x}}_{i\rightarrow0}) \\
\hat{\epsilon}_i &= \frac{\mathbf{x}_{t_i} - \sqrt{\alpha_{t_i}}\,\hat{\mathbf{x}}_{i\rightarrow0}}{\sqrt{1-\alpha_{t_i}}} \\
\mathbf{x}_{t_{i-1}} &= \sqrt{\alpha_{t_{i-1}}}\,\hat{\mathbf{x}}_{i\rightarrow0}
                      + \sqrt{1-\alpha_{t_{i-1}}}\,\hat{\epsilon}_i
\end{align}

The batch loss is the average  
$\mathcal{L} = \tfrac{1}{3}(\mathcal{L}_3 + \mathcal{L}_2 + \mathcal{L}_1)$.


\subsection{DDPM Used in Lifting}

\paragraph{DDPM Sampling.}
\label{par:DDPM_sampling}
Full-sampling quality is best with DDPM, so for lifting we apply guidance within the DDPM inference procedure:
\begin{align}
\mathbf{x}_T &\sim \mathcal{N}(0,\mathbf{I}) \\
\text{for } t = T, T-1, \ldots, 2,1&: \\
\hat{\mathbf{x}}_{0} &= D_\theta(\mathbf{x}_t,\, t) \\
\epsilon_t &\sim \mathcal{N}(0, \mathbf{I}) \\
w_{\hat{\mathbf{x}}_{0}} &= \frac{\sqrt{\alpha_{t-1}}\beta_t}{1-\alpha_t} \\
w_{\mathbf{x}_t} &= \frac{\sqrt{1-\beta_t}\,(1-\alpha_{t-1})}{1-\alpha_t} \\
w_{\epsilon_t} &= \sqrt{\frac{1-\alpha_{t-1}}{1-\alpha_t}\,\beta_t} \\
\mathbf{x}_{t-1} &= 
    w_{\hat{\mathbf{x}}_{0}}\,\hat{\mathbf{x}}_{0}
  + w_{\mathbf{x}_t}\,\mathbf{x}_t
  + w_{\epsilon_t}\,\epsilon_t
\end{align}

\paragraph{Applying 2D Guidance.} \galdel{Remove this entirely?}
To apply ray-projection guidance during DDPM sampling, we insert:
\begin{align}
\hat{\mathbf{x}}_{0} &= P_\Pi(\hat{\mathbf{x}}_{0},\, \mathbf{y})
\end{align}
after equation (19).
This procedure is a specific instance of guided diffusion introduced by \cite{song2021scorebasedgenerativemodelingstochastic} and a relaxation of \cite{li2024coin}, which handles the harder case of unknown camera.

\section{Human Preference Survey}
\label{app:human_survey}
The human surveys were conducted through a web-based interface with anonymous crowdworkers recruited via Prolific. Participants were unaware of which method generated which motion; clips were labeled only as \textit{Video~A} and \textit{Video~B} with randomized left/right assignment. Both clips played automatically on page load, and choice buttons remained disabled until both had finished playing, ensuring
each motion was watched in full. 

\begin{figure}[t]
    \centering
    \includegraphics[width=\linewidth]{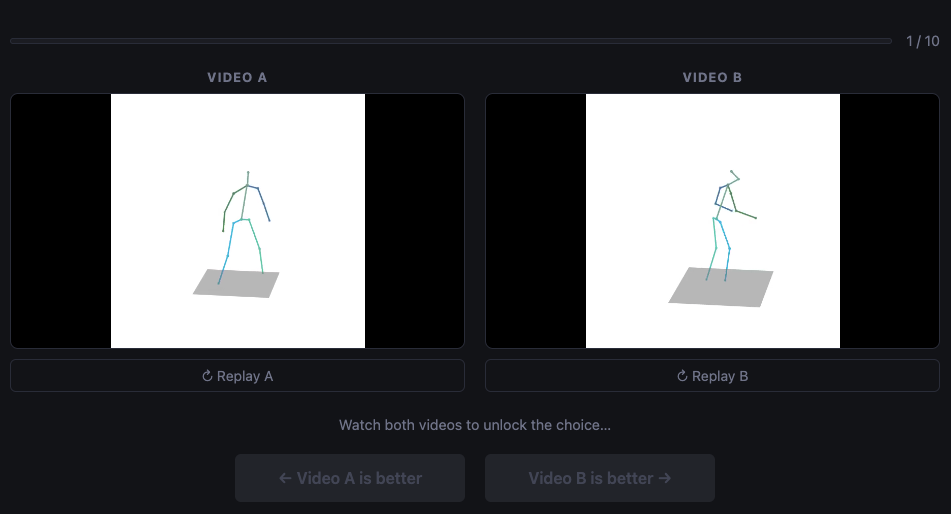}
    \caption{A representative screenshot from the NBA Human Preference Survey interface.}
    \label{fig:NBA_human_survay_screenshot}
\end{figure}

 The \textbf{NBA survey} (20 participants) compared our method against MAS~\cite{kapon2024mas} for unconditional basketball motion generation. Each participant evaluated 10 pairs of stick-figure animations and selected the one that looked more realistic and natural. No text prompt or reference motion was provided. A representative screenshot from the survey interface is shown in \cref{fig:NBA_human_survay_screenshot}.

\begin{figure}[t]
    \centering
    \includegraphics[width=\linewidth]{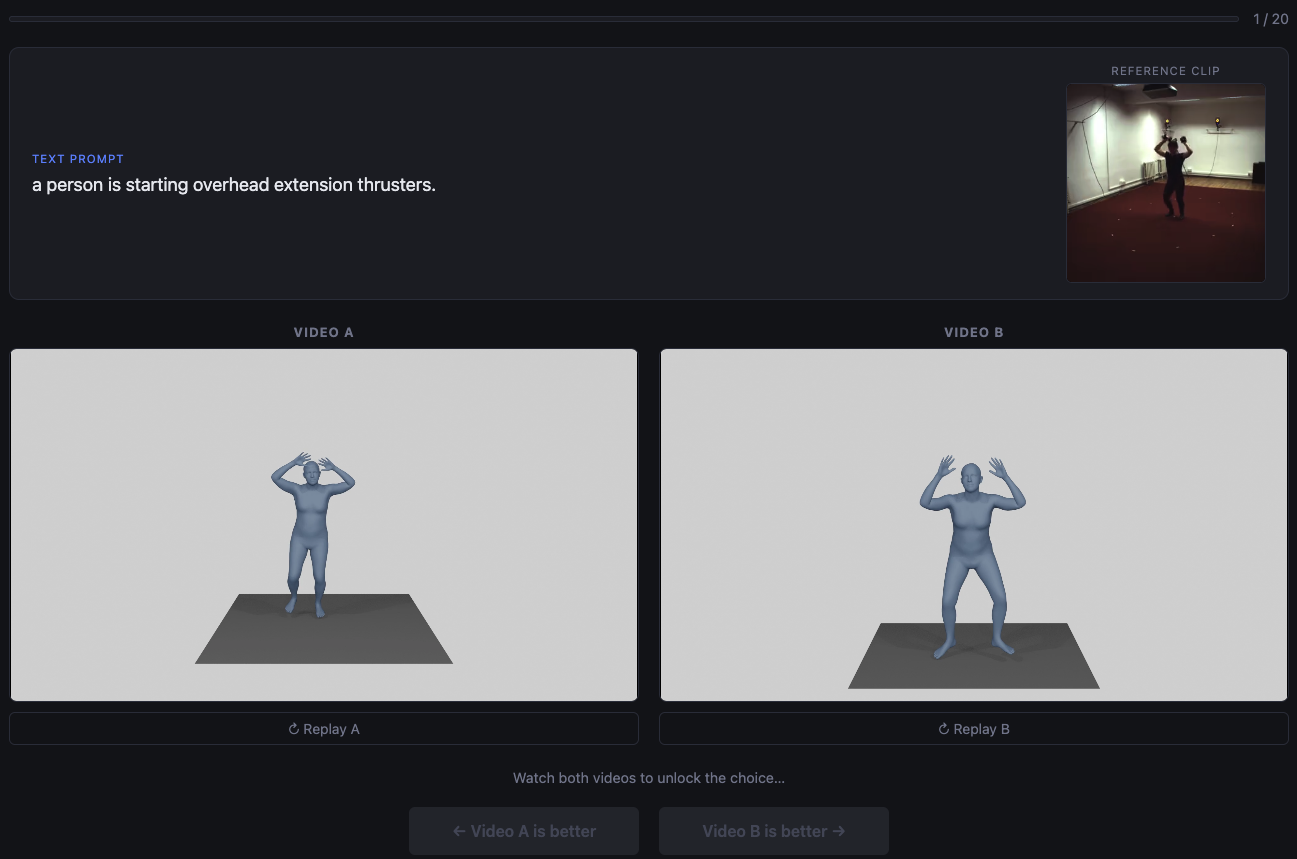}
    \caption{A representative screenshot from the Fit3D Human Preference Survey interface.}
    \label{fig:human_survay_screenshot}
\end{figure}

The \textbf{Fit3D survey} (52 participants) evaluated six methods for text-conditioned motion generation. Each participant evaluated 20 pairs, one per text prompt, where the two methods shown were drawn at random from the six candidates. Each page displayed the text prompt used for sampling and a reference video from the Fit3D dataset~\cite{Fieraru_2021_CVPR} to ground participants' understanding of the described action, alongside the two candidate clips.
A representative screenshot from the survey interface is shown in \cref{fig:human_survay_screenshot}.

 For both surveys the following text instructions were introduced:

\textbf{Participant Consent Form}

\textbf{Study title:} Human Evaluation of AI-Generated Human Motion \\
\textbf{Purpose:} This study investigates the perceptual quality of computer-generated human motion animations. \\
\textbf{Participation:} You will watch 20 pairs of short video clips and indicate which appears more realistic (~2--4 minutes). \\
\textbf{Risks:} There are no known risks. \\
\textbf{Confidentiality:} No personally identifiable information is collected. All responses are anonymous. \\
\textbf{Voluntary:} You may stop at any time without penalty. \\
\textbf{Compensation:} Upon completion you will receive a unique code to claim your compensation.

\textbf{ Motion Generation Survey}

You will evaluate \textbf{20 pairs of motion clips}. Each pair shows two animations for the same text prompt; choose whichever looks more \textbf{realistic and natural}.

\begin{itemize}
  \item Read the text prompt at the top of each page.
  \item Both videos play automatically; choice buttons unlock after both finish.
  \item You may replay clips as many times as you like before choosing.
  \item The survey takes approximately 2--4 minutes.
\end{itemize}

Participants were compensated according to the survey duration, with an average effective rate of £15.53/hour across both surveys.

\section{Camera Parameters Estimation}
\label{PNPdetails}
When camera parameters are unavailable, we estimate them by solving EPnP between the first 24 frames (minimal motion length used by MDM \cite{tevet2023human}) of the 3D estimated pose (MVLift \cite{li2025mvlift} results on HumanML3D \cite{Guo_2022_CVPR} and WHAM \cite{wham:cvpr:2024} on Fit3D \cite{Fieraru_2021_CVPR}) and the 2D poses. To do so we use OpenCV \cite{opencv_library} EPnP solver and Levenberg-Marquardt pose refinement, both with default parameters.

\section{Explicit HumanML Channel Partitioning}
\label{app:HumanML_chunnels}
HumanML3D's representation \cite{Guo_2022_CVPR} is composed of:
\begin{enumerate}
    \item 1 channel for angular velocity around the y-axis, 2 channels for root velocity in the XZ plane, 1 channel for root height.
    \item 3 channels per non-root joint, representing X (root coordinate frame) Y (global) and Z (root coordinate frame).
    \item 6 channels per non-root joint, representing the 6D continuous rotations of the joints in relation to the rest pose angle (T-shape human), each joint rotation is calculated as the normalized displacement from its ancestor.
    \item 3 channels per joint (including root) representing the per-joint velocity.
    \item 4 channels representing the 4 foot contact flags. For the NBA dataset with only 2 foot joints we replicate these flags per foot.
\end{enumerate}

So in total our $x$ is composed of $A_J=4 + (J-1)\times 3$ channels and $\mathbf{r}$ of $B_J=(J-1) \times6 + J\times3+4$ channels.

\raggedbottom

\end{document}